\documentclass[acmlarge,screen,nonacm]{acmart}

\AtBeginDocument{%
  }

\usepackage{adjustbox}
\usepackage{tabularx}
\usepackage{ragged2e}
\usepackage{url}

\usepackage{amssymb}
\usepackage{enumitem}
\usepackage{array}

\usepackage{tabularx}
\usepackage{graphicx} % Required for inserting images
\usepackage{caption}  % Required for captioning
\usepackage{subcaption} % Required for subfigures

\usepackage[utf8]{inputenc}
\usepackage{rotating} % Required for \turnbox or similar reflections
\DeclareUnicodeCharacter{01DD}{\reflectbox{e}}

\usepackage{forest}
\usepackage{xcolor}
\usetikzlibrary{shadows}
\tikzset{
        my node/.style={
            draw,
            minimum width=1cm,
            rounded corners=1,
        }
    }
\setcopyright{none}
\renewcommand\footnotetextcopyrightpermission[1]{}

\begin{document}

\title{Rethinking Indic AI from a Lens of Cultural Heritage Preservation}
%\title{Culture Sensing: A Survey of Indic NLP Foundations and a Research Agenda for AI in Cultural Heritage Preservation}
% \title{Foundations of Indic NLP and Culture Sensing: Survey and Recommendations}
%\title{Indic NLP and Cultural Connect: A Longitudinal Survey and Recommendations}
% \title{AI and Indic Linguistic Foundations: A Longitudinal Survey and Recommendations}

\author{Aparna Madva}
\email{aparna.m@iiitb.ac.in}
\orcid{0009-0004-2264-2642}

\author{Sharath Srivatsa}
\email{sharath.srivatsa@iiitb.ac.in}
\orcid{0009-0002-9404-8250}

\author{Srinath Srinivasa}
\email{sri@iiitb.ac.in}
\orcid{0000-0001-9588-6550}

\author{Tulika Saha}
\email{Tulika.saha@iiitb.ac.in}
\orcid{0000-0002-3252-0997}

\affiliation{%
  \institution{International Institute of Information Technology, Bengaluru}
  \city{Bengaluru}
  \state{Karnataka}
  \country{India}
}


\begin{abstract}
As Artificial Intelligence (AI) makes inroads into different parts of the Indian subcontinent, there is significant interest in studying how AI impacts the linguistic and cultural foundations of this civilization. AI is seen as a ``double-edged sword'' where on the one hand, it can enable access and inclusion for a large population, on the other, it can homogenize worldviews and exclude underrepresented languages and worldviews. In this paper, we try to characterize this problem by addressing the extensive characteristic nature of Indian linguistics and the way they closely connect to cultural practices and worldview. We then perform a longitudinal survey of how Natural Language Processing (NLP) techniques have evolved in this space, tracing the historical development of Indic NLP, covering key milestones, methodological shifts, and resource creation efforts. In addition, the paper also examines the structural and sociolinguistic characteristics of Indian languages, such as rich morphology, complex scripts and grammar rules, diglossia, and large dialectal variation, and explains how these create unique challenges for building AI foundation models. We then discuss the growing role of Indic foundation models and analyze how these models address these long-standing resource and representation gaps. Finally, we propose a research direction called `\textit{Culture Sensing}', which re-imagines AI based on hermeneutic reasoning. Culture Sensing aims to address open problems such as ensuring equitable performance across low-resource languages and producing outputs that are culturally meaningful. By bringing together past work, current techniques, and emerging trends, this paper outlines research directions that can guide the next phase of Indic NLP and contribute to the development of more robust and inclusive Indic foundation models.
\end{abstract}





\maketitle

\section{Introduction}
Indic languages refer to the wide spectrum of languages spoken in the Indian subcontinent, including India, Nepal, Sri Lanka, Pakistan, Bhutan, Bangladesh, and other countries-- that have a long linguistic and cultural history,  collectively contributing to more than a fifth of the total world population. The Indian subcontinent has a long and rich linguistic history, hosting a wide variety of languages. Each language often uses a different script and has a rich body of literature. Just in India, the constitution recognizes 22 major literary languages as official languages of the country. In addition to this, there are about 121 other non-official, but major languages, and more than 19,000 minor languages, dialects, and creoles. Languages are also culturally significant across different regions of the subcontinent, often shaping their own unique regional identity. 

\begin{figure}
    \centering
    \includegraphics[width=0.5\linewidth]{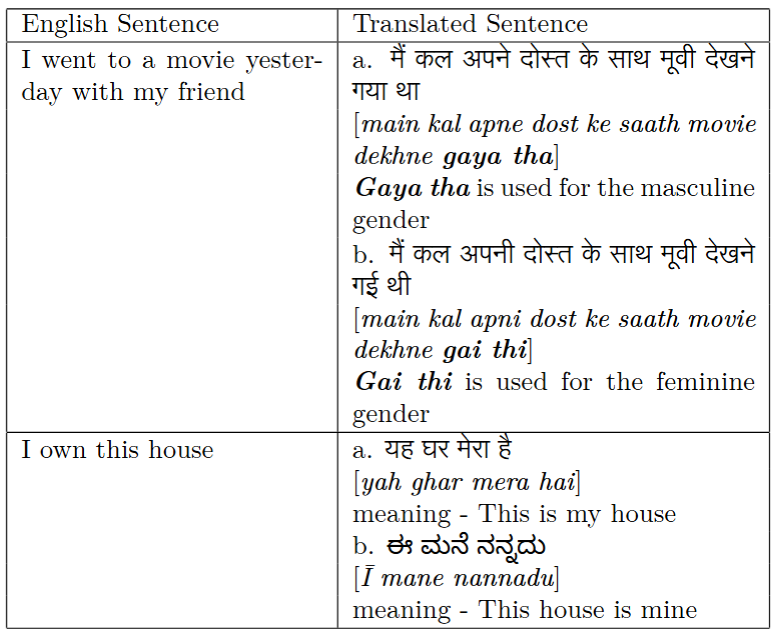}
    \caption{Translated Sentence Pairs Demonstrate the Innate Worldview of the Indic Subcontinent}
    \label{fig:translated}
\end{figure}

An innate hermeneutic diversity can be observed in the case of the Indian subcontinent. The population of the Indic subcontinent often holds diverse perspectives drawn from the sociocultural discourse. Figure~\ref{fig:translated} shows pairs of sentences, a source English sentence, and corresponding translations in the Indic languages Hindi and Kannada. In the first sentence, it can be seen that mentioning of the gender is inevitable while making a statement such as `\textit{I went to a movie yesterday with my friend}'. From the second sentence, it can be observed that words that signify `identity' rather than `ownership' are used to convey the meaning of owning a house. Such minute details not only reflect the linguistic divergence but also the underlying hermeneutic differences.

% Motivate the problem about how AI can affect this cultural subcontinent. Cite specific sources and specific schools of thought
The integration of AI into the cultural and linguistic landscape of the Indian subcontinent is bound to significantly shape the development and progress of the region. For India, AI adaptation is driven by a dual necessity: the scope to leapfrog traditional infrastructure deficits~\cite{AIforALL, 10.1145/3278721.3278738} and the desire to preserve, scale, and disseminate its immense cultural~\cite{choudhary2025leveraging} and linguistic~\cite{Krishna2025ImpactOA} diversity. AI is projected to foster inclusion, economic growth, and also social good~\cite{maheshwariai}. India's diverse sociocultural fabric can also benefit from using AI for the restoration and preservation of culture and heritage. It can enhance traditional values like \textit{Vasudhaiva} \textit{Kutumbakam} (meaning - the world is one family), ensuring focus on inclusive, community-centric technology. 

Apart from this, improvement in digital literacy and usage has resulted in an increase in Indian language content on the Internet across multiple platforms, including social media, news media, and others. Technological advances have been penetrating quickly into the Indian population, with the number of Indian language Internet users estimated to exceed 900 million as of the year 2025~\cite{indian-languages-internet}. This advancement calls for language technology that enables people to access information in their native language.

However, the potential of AI is faced with significant bottlenecks for adoption, primarily due to the digital divide and linguistic challenges. This is predominantly observed in the current-day Large Language Models (LLMs), that are prone to \textit{homogenization} of hermeneutic interpretations due to lopsided representations of disparate worldviews in their training data, as well as lopsided representation of user base and their feedback that fine-tunes the model's responses over time. The issue is amplified in the case of Indic languages since LLMs are shown to align disproportionately with the linguistic patterns of specific subpopulations~\cite{sourati2025shrinking}. It has also been observed that an individual (or groups) exclusively receive negative outcomes in algorithmic decision making. For example, a job applicant may get rejected from a lot of job opportunities they apply to when all the companies make use of similar resume screening algorithms~\cite{NEURIPS2022_17a234c9}.  The major reason for algorithmic bias is the lack of representation for not just Indic language text, but also the diverse worldviews in the Indian subcontinent. A majority of the LLMs are trained on translations from English language datasets that are collected mostly in urban contexts. When applied to Indic contexts, they fail to generalize to the different cultural nuances. 

Multiple initiatives have been focusing on Natural Language Processing for Indic languages-- or Indic NLP for short, with the goal to build representative datasets in multiple Indic languages, and also to model the different linguistic features of Indic languages. Indic NLP  has its own evolutionary history, somewhat paralleling mainstream research in NLP that is most prominently focused on English, and to a lesser extent, on other European languages. A considerable amount of research focusing on language technology in Indian languages has been conducted for a long time and is continuously pursued at present. Research on cultural alignment of LLMs~\cite{10.1145/3772318.3790519, agarwal2026fluentforeignregionalllms} is also gaining traction. These studies have contributed valuable resources, such as a corpus and models, to the Indic NLP community. 

It is important to ensure that both the hermeneutic and linguistic diversity is cohesively preserved. While there has been substantial development in Indic NLP pertaining to models, datasets, and benchmarks, representation of the multitude of worldviews is still lacking in current-day language models. To preserve the cultural diversity in the Indian subcontinent, it is essential to ensure that the corresponding hermeneutic plurality is represented in the models being built.

In this paper, we present a longitudinal study of research on NLP for Indic languages. We provide an overview of the evolution of research in this area and the different paradigms explored. Due to the dynamically changing NLP landscape, we include studies published only till 2025 to the best of our knowledge. We also investigate prevailing challenges, including the attainment of equitable performance across low-resource languages and dialects, and the preservation of cultural and linguistic fidelity in generative outputs. 

We also introduce a new research direction, \textit{Culture Sensing}, that rethinks the framework of AI based on hermeneutic reasoning. We substantiate this idea with the help of two different use cases where AI techniques are utilized to analyze lesser-known knowledge systems. Culture Sensing utilizes the vibrant multimodal knowledge systems in various native communities and represents the corresponding worldview in AI models. By this, the focus is on making AI models pluralistic and inclusive, thereby preserving lesser-known knowledge systems on the brink of extinction. By bringing together foundational research, contemporary methodologies, and emerging trends, this paper sets out strategic research directions intended to facilitate the advancement of robust, culturally inclusive Indic NLP. 

Section~\ref{sec:characteristics} provides the background of developments in the NLP field and distinguishing features of Indian languages to be considered for building efficient NLP systems. Section~\ref{sec:history} reviews different Indic NLP works since their inception till the recent breakthroughs, considering the evolving approaches. Section~\ref{sec:challenges} provides an overview of unique challenges for IndicNLP and discusses mitigation strategies. Section~\ref{sec:language} discusses the proposed idea of Culture Sensing to motivate the need for culturally inclusive AI and provides directions towards its achievement.

% For building language models specific to Indic languages, it is important to understand a few of the unique features of Indian languages. Indic languages are phonetic (meaning they are pronounced as they are written), and have a complex morphological structure. Agglutination is frequently used, where the role marker for a word in a sentence is attached to it as a suffix and enables free word order sentences. Due to this, a model architecture that is successful in the case of the English language might not adapt very effectively to Indic languages.

\section{Characteristics of Indic Languages}
\label{sec:characteristics} 
Languages in the Indian subcontinent have a rich linguistic heritage and a literary tradition that spans across millennia. Words in Indic languages have complex morphology, and the grammar for their organization includes elaborate rules. An understanding of the distinctive features of Indic languages is necessary for building meaningful language technology for them. This section explores the basic structure and essential features of Indic languages along with examples.

\begin{figure}[h]
    \centering
    \begin{forest}
        for tree={%
            my node,
            l sep+=5pt,
            edge={gray, thick},
        }
        [Features of Indic Languages, fill=blue!20
        [Morphology, fill=pink
        [\textit{Akshara} System, fill=pink]]
        [Grammar, fill=purple!50
        [Free Word Order, fill=purple!30][Agglutination, fill=purple!30[Inflection, fill=purple!10][Derivation, fill=purple!10][Compounding, fill=purple!10] ][\textit{Sandhi}, fill=purple!30][\textit{Kaaraka}, fill=purple!30]]
        [Diglossia, fill=yellow!50
        [Regional, fill=yellow!30][Communal, fill=yellow!30]]
        ]
    \end{forest}
    \caption{Characteristics of Indic Languages}
    \label{fig:features}
\end{figure}
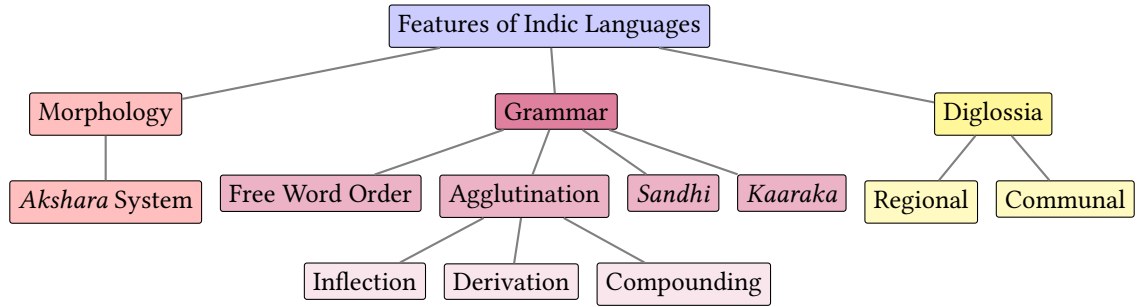

\subsection{\textit{Akshara} System of Indian Languages}

\begin{figure}
    \centering
    \begin{subfigure}[b]{0.6\textwidth}
        \includegraphics[width=\textwidth]{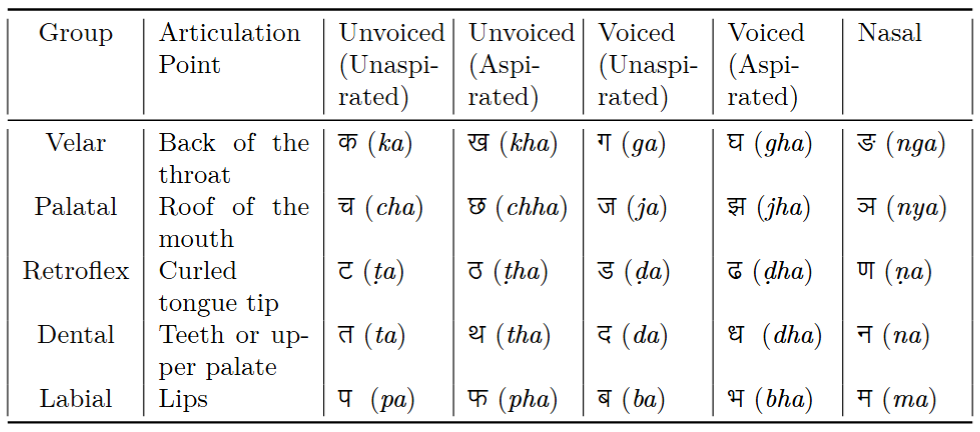}
        \caption{Structured Consonants}
        \label{fig:s_consonants}
    \end{subfigure}
    \par\bigskip
    \begin{subfigure}[b]{0.6\textwidth}
        \centering
        \includegraphics[width=\textwidth]{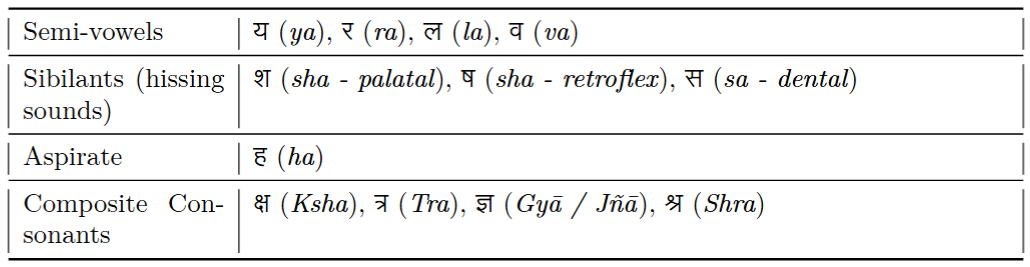}
        \caption{Unstructured Consonants}
        \label{fig:u_consonants}
    \end{subfigure}
    \caption{Consonants in Indic Languages}
    \label{fig:consonants}
\end{figure}

Indic languages are predominantly phonetic in nature, where the alphabet is based on observing variegated forms of sounds emanating from the human vocal system comprising of throat, tongue, lips and nasal elements. The alphabet of a majority of Indic languages contains 33 consonants and 15 vowels. Some languages can also include additional 1-2 consonants and vowels. The consonants contain an inherent vowel and are usually divided into 25 structured consonants (shown in Figure~\ref{fig:s_consonants}) and the remaining unstructured consonants (shown in Figure~\ref{fig:u_consonants}). Consonants are further clustered into different segments depending on which part of the human vocal system is used to produce that class of sounds. 

An individual letter of the alphabet is called an \textit{akshara} that consists of a vowel and 0 or more consonants. Each akshara has its own form and sound. The letters and words are pronounced exactly as they are written, since the languages are phonetic. The number of written symbols can be more than the number of characters in the alphabet, as more than one consonant can be combined without an intervening vowel to form digraphs. Unlike English, where sounds are constructed by lexical concatenation of letters from a base alphabet, Indian languages construct syllables by modifying consonants with verbs and other consonants.

\begin{figure}[h]
    \centering
    \begin{subfigure}[b]{0.35\textwidth}
        \centering
        \includegraphics[width=\textwidth]{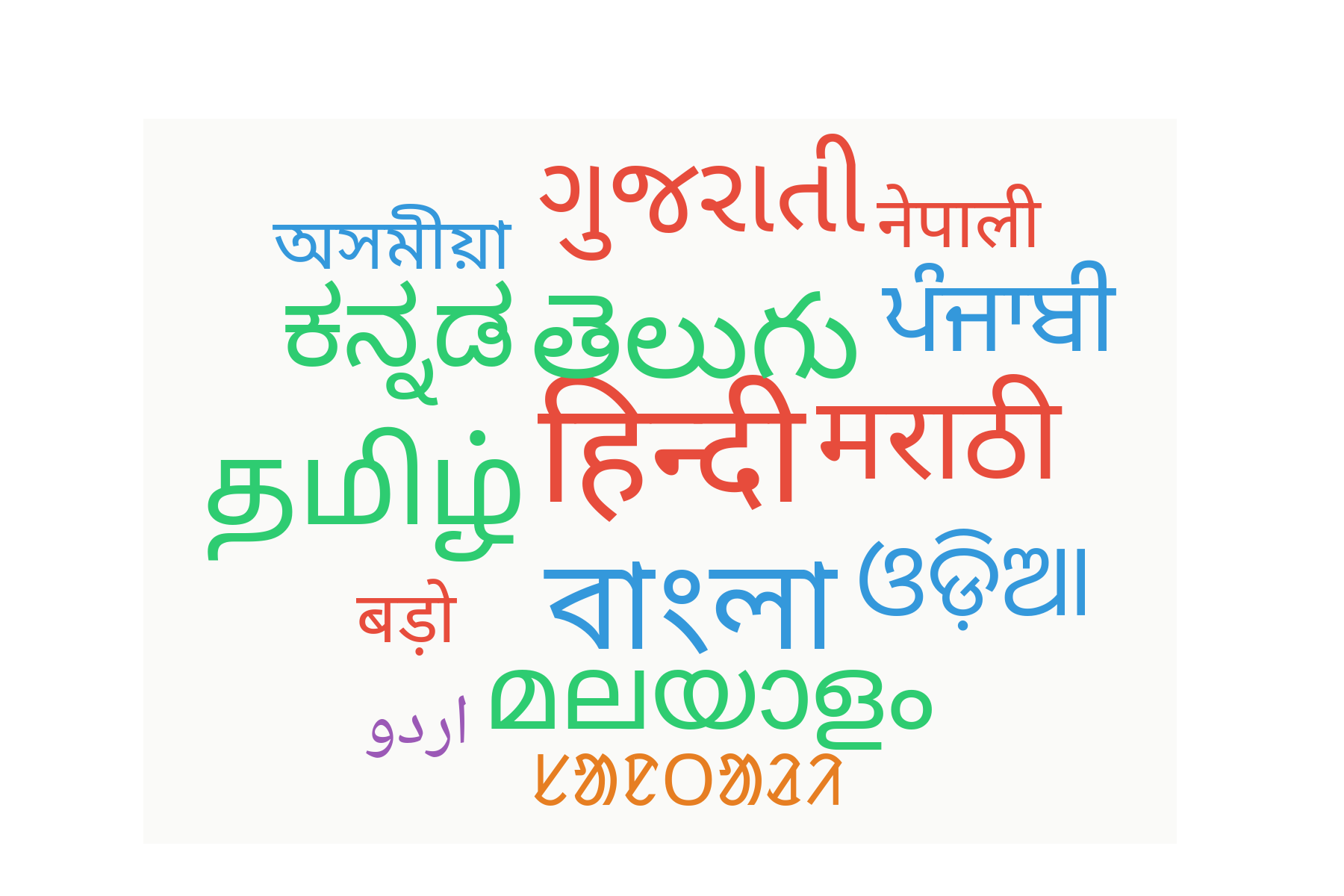}
        \caption{Different Indic Language Scripts}
        \label{fig:scripts}
    \end{subfigure}
    \hfill
    \begin{subfigure}[b]{0.55\textwidth}
        \centering
        \includegraphics[width=\textwidth]{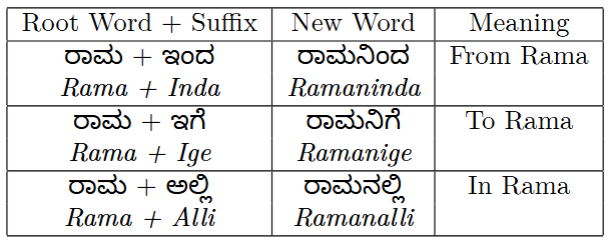}
        \caption{Agglutination Forming Different Words With One Root Word}
        \label{fig:diacritic}
    \end{subfigure}
    \caption{}
    \label{fig:features}
\end{figure}

\begin{figure}[h]
    \centering
    \begin{subfigure}[b]{0.35\textwidth}
        \centering
        \includegraphics[width=\linewidth]{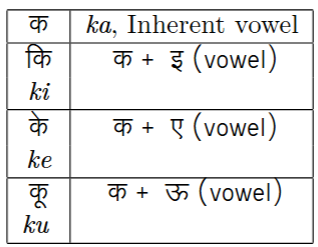}
        \caption{Letter Formation with Consonants and Vowels}
        \label{fig:vowels}
    \end{subfigure}
    \hfill
    \begin{subfigure}[b]{0.55\textwidth}
        \centering
        \includegraphics[width=\linewidth]{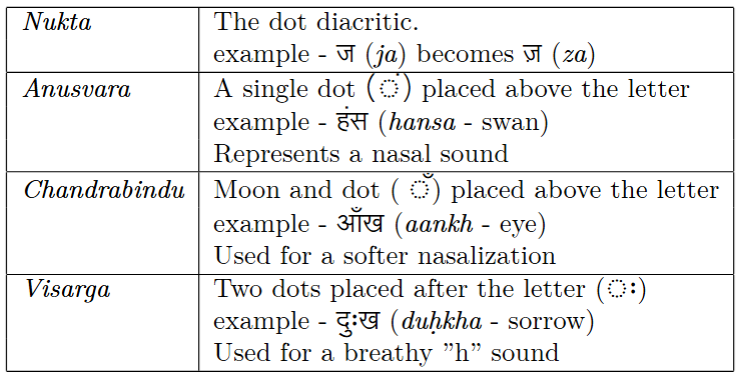}
        \caption{Different Diacritics used in Indic languages}
        \label{fig:diacritic}
    \end{subfigure}
    \caption{Inherent Features of Indic Languages}
    \label{fig:features}
\end{figure}

Even though the alphabets of Indic languages are similar, the same is not reflected in the scripts used to write them. The written script has wide variability for different languages in India, as shown in Figure~\ref{fig:scripts}. Differences can be observed even in the structure of graphemes and their combinations. The rules for the formation of graphemes vary within a script and also across scripts. For example, the Devanagari script used to write Hindi language includes a horizontal line called \textit{Shirorekha} at the top of the word that connects every character of the word. But this style of writing is not used by every Indic language, and a few of them write graphemes without touching each other. The alpha syllabic writing system is used, where consonants and vowels are represented as a single unit, as shown in Figure~\ref{fig:vowels}. Consonants represent a letter, while vowels can either appear in atomic form, not connected to consonants, or are dependent on the consonant and marked like a diacritic or some other form of modification. A vowel can appear to the left, right, top, or bottom of the consonant. A grapheme representing a combination of vowels is also represented similarly, and the supporting consonant can be a modified form of its original form, or it can be completely different as well. A large number of diacritics are also present, shown in Figure~\ref{fig:diacritic}.

\subsection{Formalism of Indian Languages: Panini's Framework}
% The following are the main categories of morphological processes :
%\begin{enumerate}
%  \item Inflection: Produce different grammatical variants of the stem word with suffixes, where stem words can be a noun or a verb, and suffixes represent tense, number, case, or gender. The case of the stem word is unchanged. For example, the Hindi language word \textit{gaadiwala} (carriage driver) in plural form is written as \textit{gaadiwale}, with a modified suffix
%  \item Derivation: Suffixes are added to a stem word to create a new word with additional or modified meaning. The case of the stem word can change. For example, the Kannada language word \textit{chitra} (picture) when suffixed with \textit{kaara} (doer) form the new word \textit{chitrakaara} (artist)
%  \item Compounding:  Forming complex word structures by combining two or more free morphemes
%\end{enumerate}

The formal structures of most Indic languages are either directly based on, or are greatly influenced by the framework established by Panini, called Astadhyayi, written around the $6^{th}$ century BCE~\cite{bharati1993parsing}. This framework of structuring language also influences the way language is communicated and characteristics the resultant hermeneutic structures. In this section, we look at some key elements of the Paninian framework that are seen in almost all Indic languages. 

\textit{Morphemes} are elemental meaning-bearing words in Indic languages, that can in turn, be considered to have two elements: the \textit{stem} and the \textit{affix}. Prefixes, suffixes, circumfuse, and infixes are different types of affixes. Prefixes precede the stem word, suffixes follow the stem word, and circumfuse morphemes precede and follow the stem word. Affixes inserted into stem words are called Infixes. Stem words carry the primary meaning, and affixes give various kinds of additional meaning to the stem word~\cite{neupane2024morphological}.

%Understanding the morphology of a language is essential to performing computational linguistics or NLP on it. Panini's Ashtadhyayi or Paninian Framework delves into the formalism of the morphology of the Sanskrit language prevalent in his times, which is predominantly applicable to most Indian languages \cite{bharati1993parsing}. This work categorizes words into the four types\footnote{\hyperlink{https://ashtadhyayi.com/}{https://ashtadhyayi.com/}} shown in Figure~\ref{fig:ashtadhyayi}. Subsequent paragraphs detail more information from Panini's Framework, widely applicable to Indic languages.

%\begin{figure}
%    \centering
%    \includegraphics[width=0.9\linewidth]{images/ashtadhyayi.png}
%    \caption{Four Word Categories according to %\textit{Ashtadhyayi}}
%    \label{fig:ashtadhyayi}
%\end{figure}

%A \textit{upasarga} is a word or affix that is added at a word's beginning to change or modify its meaning. Occasionally, \textit{upasarga} also refers to a set of twenty prepositional prefixes added to verbs or action nouns. For example, the prefix \textit{vi} (apart) added to the word \textit{desh} (country) results in the word \textit{videsh} (foreign country) in Hindi. 

Among the affixes are a set of case modifiers that are used to assign roles to specific words. This is enabled with the concept of \textit{kaaraka} to depict the role of a given word in a sentence. A set of suffixes called \textit{vibhakti} is usually used to express \textit{kaaraka}. The number of \textit{vibhakti}s in a language can be 7 or 8. This concept can be related to the idea of grammatical cases in languages like English and German. For example, the word \textit{raja} (king) can be added with the suffix \textit{annu} that represents \textit{dwitiya} vibhakti to form the word \textit{rajanannu} and the corresponding \textit{kaaraka} as \textit{karma kaaraka}, which refers to the object of an action. Hence, the resultant word \textit{rajanannu} forms the object of a sentence it is part of. These role modifiers give the property of \textit{free word order} to Indic languages, where the position of role-modified words in a sentence can be changed without modifying the overall meaning of the sentence. The concept of \textit{vibhakti} has to be considered while designing for tasks such as Named Entity Recognition (NER) in Indic languages. It has been observed that the identification of an entity may depend on its suffix corresponding to an appropriate \textit{kaaraka} in the case of the NER task for the Kannada language~\cite{doi:10.36227/techrxiv.24580582.v1}.
%\begin{figure}
%    \centering
%    \includegraphics[width=0.9\linewidth]{ACM_Journals_Primary_Article_Template/freewordorder.png}
%    \caption{Free word order translations}
%    \label{fig:freewordorder}
%\end{figure}

%The free word order nature of Indic languages can sometimes pose challenges in neural machine translation where predicting a masked word may favor most commonly used word orders. Figure~\ref{fig:freewordorder} shows an example of four sentences translated from Kannada to English using Google translate. While the first two sentences are word permutations of the same sentence, the last two are different sentences with a slightly different meaning. But all of them are translated to the same sentence. 

Another concept in Indic languages that gives it an agglutinative nature is \textit{Sandhi} or morphophonemic rules that allow a word to be joined with other words or morphemes, resulting in a newly formed word with phonological changes and also possible orthographic variations. For example, joining the Kannada words \textit{mane} (house) and \textit{inda} (from) results in \textit{maneyinda} (from house) following the \textit{Agama sandhi}~\cite{zydenbos2020manual} rule. Compounding or \textit{samaasa} is also observed frequently~\cite{compounding}. Here, two or more meaningful words are joined together to output a (longer) word such that its meaning is derived from the underlying words. The produced word contains more than one stem. For example, the Sanskrit word \textit{vidyanipuna}, meaning one who is sharp in the studies, is formed by two words \textit{vidya} (education) and \textit{nipuna} (expert), following the \textit{Tatpurusha samasa} rule. These phenomena not only obscure word boundaries, but also modify the characters at the joining point of the words. This makes the task of word segmentation and tokenization complex in the case of Indic languages. Hybrid approaches for word tokenization, such as the one proposed by~\cite{sandhan2022translisttransformerbasedlinguisticallyinformed}, have demonstrated improved performance using knowledge of the sandhi phenomenon.

Bharati and Sangal~\cite{bharati1993parsing} describe a computational grammar formalism for Indian languages in general from Paninian Grammar, which was composed for Sanskrit, with the idea that Sanskrit can form the bridge language for translating between any two Indian languages. The authors show a compact formalism to parse simple and complex sentences and active and passive voices in sentences using the invariants appearing in verbal and noun parts of the sentence, while the sentence, on the whole, can follow free word order. The notion of Kaaraka relations is the core of the described Panian Framework. Kaaraka relations are syntactico-semantic (or semantico-syntactic) relations that link verbal parts of the sentences to noun parts by invariants appearing with root verbs and nouns in a free-word order sentence. The invariants in noun parts are vibhaktis or postpositions, and the invariants of verbal parts are tense-aspect-modality (TAM) elements. The invariants help capture the sentence's gross meaning even when word order is changed. Moreover, a change in word order conveys secondary information like emphasis, etc., and does not affect the primary meaning of the sentence. The verbal group invariant forms impose or demand specific invariants in the noun group, and hence, the verbal group is the demand group, and the noun group is the source group. The invariants in demand and source are mainly prevalent in the spoken and written texts of Indian languages, which makes the Paninian Framework a suitable formalism for Indian languages. 

There are two components to establishing the kaaraka relations. The default Kaaraka chart, where three types of kaarakas or nominals, Karta, Karma, and Karana, are mapped to post positions or vibhaktis, and used to establish a relation when the verb group has basic TAM. The other map is the transformation rules, which map all possible TAMs in the verbal group to specific postpositions in the default kaaraka chart. One can use the kaaraka relation formalism to build a parser. The first stage is morphology analysis of a sentence, which involves obtaining grammatical information about words from a lexicon or dictionary and local word grouping of noun groups and verbal groups. Noun groups contain postposition markers with nouns, and verbal groups contain verbs and their auxiliaries. 

% \cite{bharati1991local} details the rules of local word grouping. The core parser identifies kaaraka relations and the sense of the words with the local word group. For each verbal group based on the verb, the system loads kaaraka charts. We can view each entry in the kaaraka chart as a restriction on the noun group, which has to be satisfied in the sentence. Kaaraka restrictions have optional and mandatory restrictions. We construct a constraint graph using the kaaraka relations. When a kaaraka restriction is satisfied, an arc will be created from the demand or verbal group to the source or noun group. A subgraph in the constraint graph is a valid parse if there is precisely one arc from demand to source group for a mandatory restriction or, at most, one arc from demand to source group for an optional restriction, and only one incoming arc to a source group. If several subgraphs satisfy the above conditions, then the sentence is ambiguous, and if no subgraph satisfies the above restrictions, the sentence is probably ill-formed. In contrast to widely used dependency parsing, the Paninian Framework uses linguistic insights to build kaaraka charts and transformation rules, and restrictions ensure bipartite graphs. The constraint graph of a sentence constructed through kaaraka charts enables parsing with well-known graph parsing approaches. 

\subsection{Diglossia in Indian languages}
Diglossia refers to the difference in the way in which a language is spoken and used in colloquial settings as against the codified, formal variety that is often used in formal education~\cite{diglossia}. This becomes an essential consideration while building language models since the training data for language models mostly belongs to the formal language variety. 

In a majority of Indic languages, the spoken (colloquial) style of the language differs significantly from the literary (formal) variety. Multiple dialects of a language exist, and these dialects can also exhibit distinguishing characteristics. Among multiple dialects, there can be differences in terms of phonology, morphology, lexicon, and syntax. The distinguishing factor for dialects can be regional or social. In some cases, the words can be entirely different between dialects. For example, the word for money is \textit{duddu} in Bengaluru Kannada, whereas it is \textit{rokka} in Dharwad Kannada. Similarly, some dialects of Hindi are Khari Boli, Braj, Bundeli, Marwari, Kumauni, and Garhwali. Some dialects of languages are also characterized by the community speaking it. For example, Soliga Kannada is used by the Soliga tribe and has some Tamil influence. Halakki Kannada is spoken by a tribe called Halakki Vokkaligas. A few members of the Gowda community living in the Kodagu district of Karnataka use the Arebhashe language. In addition to this, the inter-generational tacit knowledge is communicated orally using colloquial style languages, especially in the communities in rural areas. All of these variations are synchronic, and they occur within a single point in time. 

There also exist diachronic variations, when languages evolve through multiple historical periods. For example, Kannada language evolved through three periods: \textit{Halegannada}, \textit{Nadugannada}, and \textit{Hosagannada}. \textit{Halegannada} nurtured classical literature, and most of the popular literature belongs to either \textit{Nadugannada} or \textit{Hosagannada}. Many phonological variations can be observed between these language varieties. LLMs can be efficiently used to quantify the diachronic changes as demonstrated by~\cite{hariharan2025transformerenableddiachronicanalysisvedic}. There is a need for detailed research in this area.

Language models are generally trained using a corpus that uses a literary style of language. These language models hence contextually understand the formal style of the language. However, it's not necessary that the same model demonstrates similarly when its input is with some dialect or colloquial variety of the same language. Building models that generalize for dialects is also challenging, as shown for the task of speech recognition in ~\cite{dialects}, and needs better strategies. An ideal language model should be inclusive and demonstrate competitive performance for different dialects of the same language.

\section{Historical Context: Evolution of Indic NLP}
\label{sec:history}

Different paradigms of NLP have been utilized for processing Indic language data and for building models for Indic languages over time. The dynamically changing NLP landscape, right from rule-based processing techniques that used the structure of these languages to form processing rules to the recent transformer architecture, has been experimented with large quantities of data and evaluated for their performance in the case of Indic languages. We cover the research works on IndicNLP through the different techniques to the best of our knowledge in the following subsections.

\subsection{Rule-based Indic NLP}

Rule-based NLP is an approach for processing text using manually crafted, predefined linguistic rules and patterns. It relies on handcrafted rules built using expert-defined grammar and dictionaries, and pattern matching using regular expressions to analyze the syntax and semantics. Although this approach was highly accurate for specific tasks, it was time-consuming,  provided limited generalization, and was rigid. Early rule-based IndicNLP works mostly focused on  Machine Translation (MT), parsing the Indic language text, and building lexical databases such as WordNet for Indian languages. Table~\ref{tab:rule-based} provides a comprehensive list of rule-based Indic NLP models and the tasks they addressed.

\subsubsection{Machine Translation}
The MT systems performed Indic-English, English-Indic, and Indic-Indic translation. Three types of strategies were used for the development of MT systems: 

\begin{itemize}
    \item Direct translation
    \item Interlingua strategy
    \item Transfer strategy
\end{itemize}

The direct translation strategy relied on resources using dictionaries, morphological analysis, and text processing systems for translation involving a specific source and target pair of languages without any intermediate representation. The transfer strategy first performed text analysis according to the structure of the source language. Following this, processing in the context of the target language was carried out to perform 'transfer' at the lexical level, syntactic level, or at the semantic level. A target language independent universal representation was used by the Interlingua strategy for a given source language text. Any ambiguities were presumed to be resolved in this representation, and hence it should be usable to generate text in any target language.

Anglabharti, an English-Indian language translation system, was proposed by~\cite{ANGLABHARTI}, which used a rule-based system using patterns with Context Free Grammar (CFG) like structure for the source language (English). With Hindi as the target language, a prototype system with about 50 rules for English to Hindi translation was reported to be able to translate most of the common sentences. Another machine translation system, Anusaaraka, proposed by \cite{bharati2003anusaarakaovercominglanguagebarrier} was able to perform translation for the Telugu, Kannada, Marathi, Bengali, and Punjabi to Hindi using human assistance. For preserving the information while translating from a source language to a target language, substitutibility and reversibility were maintained in the translated strings as explained by ~\cite{bharati2003languageaccessinformationbased}. \cite{rao2000practical} proposed a framework for the syntactic transfer of compound complex sentences from English to Hindi using a transfer-based Machine Assisted Translation (MAT) system. Due to the inflectional nature of Hindi, this work used strategies for inflecting appropriate inflections and case markers as discussed in~\cite{rao1998natural}. An interlingua-based approach for English to Hindi translation was proposed by~\cite{dave2001interlingua}, using Universal Networking Language (UNL) interlingua. This work investigates the ability of an interlingua-based approach for translation in handling cross-language divergences.

\subsubsection{Parsing}
In case of parsing, it is observed that the dependency parsing framework is better suited for free-word order languages, including Indic languages~\cite{Bharati95}. This approach was elaborated in an early study by \cite{bharati-sangal-1990-karaka} in 1990, which proposed an overall parsing strategy for Indian languages using the \textit{kaaraka} relations described in Paninian grammar. This work proposed a parsing system with three components: a morphological analyzer to extract roots of each word in the input sentence, along with other corresponding grammatical information, followed by a Local Word Grouper (LWG)~\cite{bharati1991local} that formed word groups using information based on adjoining words to reduce the complexity for the core parser~\cite{bharati1993parsing}. Dependency parsing analyses the words via the dependency relation between them rather than structuring the phrases hierarchically. Because of this, dependency analysis is more suited for Indian languages than CFG. \cite{begum-etal-2008-dependency} came up with an approach to perform dependency annotation on Indian languages following the Paninian framework. This study utilized the concept of \textit{kaaraka} as they can be identified syntactically and can be used to understand the underlying semantics.

An improved performance using data-driven dependency parsing was obtained by \cite{Husain2009data} following a modular cascaded approach in the case of the Hindi language. In this work, each intermediate layer or module of the parser produced a linguistically valid partial parse. The final parse obtained using this method aimed at minimizing the adverse effects of long-range dependency and the non-projective nature in free word order languages. The two-stage approach for parsing was also used by \cite{bharati2009constraint} and \cite{bharati2009two} to propose a constraint-based hybrid approach for dependency parsing in free word order languages such as Hindi. This work incorporated hard and soft constraints (H-constraints and S-constraints) in the two-stage parsing approach. H-constraints, satisfied by any grammatical sentence, include lexical and structural knowledge of the language, such as structural constraints, lexicon, and other language-specific rules. S-constraints, on the other hand, were used as preferences that can be broken by a sentence. \cite{bharati2002anncorra} proposed AnnCorra, a dependency-based tagging scheme for annotating corpora in Indian languages based on the Paninian grammatical model. The proposed generalized tag scheme aimed at efficiently annotating a corpus and developing treebanks accounting for the distinct syntactic structures in all Indian languages using a tagging scheme made up of notations, defaults, and tagsets. \cite{Kanuparthi2012derivational} proposed an algorithm for derivational morphological analysis of the Hindi language. This work utilized an existing inflectional morphological analyzer (\cite{Bharati95}) for this work. They used a list of manually extracted derivational suffixes, which were used to form a set of rules for the analyzer, along with a list of words extracted from Hindi Wikipedia. These components were used by the proposed algorithm to output both inflectional analysis and derivational analysis for a given word.

\subsubsection{Lexical Resource Creation}
A rule-based automatic annotation approach to annotate a Hindi Treebank using the Paninian dependency framework was proposed by~\cite{gupta2008rule}. This work extended the experiments in~\cite{begum-etal-2008-dependency}. Here, dependency labels were marked only for inter-chunk relations where a chunk is a set of adjacent words. The robust formulation of rules for the automatic annotator anticipated a reduction in time and effort for manual annotators in dealing with a large corpus. A comparison of the annotator with a constraint-based parser explained in~\cite{bharati2002constraint} showed that the annotator performed better than the parser for most of the dependency relations. It was also observed that the annotator helped in improving the results of the first parse of the parser and hence could also be used for post-processing. \cite{bharati2009simple} extended this work to include intra-chunk annotation and included additional linguistic features for better identification of relations. The parser proposed in this work is a language-independent engine that takes a rule file for a specific language, as mentioned by~\cite{gupta2008rule}. 

A subset of the dependency treebank proposed by \cite{begum-etal-2008-dependency} was utilized by \cite{bharati2008two} to understand a few of the crucial cues of a language, which are essential for building a robust parser in the case of the Hindi language. Among the two techniques of dependency parsing, grammar-driven dependency parsing that rules out those parses that don't satisfy the established constraints and data-driven dependency parsing that learns a probabilistic model for parsing, this work experimented with two different data-driven parsers. It was observed that the extended feature set for the parser with vibhakti labels increased the performance (here, vibhakti is a generic term for preposition, post-position, and suffix). AnnCorra and two more efforts, TransLexGram and Shabda-Sutra, aimed at creating lexical resources in Indian languages, were discussed by~\cite{bharati2003leril}. Transfer lexicon and grammar, or TransLexGram in short, attempted to produce a transfer lexicon and grammar from English to Hindi. This work generated bilingual dictionaries, parallel corpora for English-to-Indian-language translation, and simple transfer patterns that can be used directly by machine translation systems. Shabda Sutra explored the various semantic meanings of a polysemous word. \cite{kulkarni2010introducing} proposed a WordNet for Sanskrit. The WordNet contained verbs in their root forms, following the technique explained in \cite{kulkarni2009verbs} to efficiently store morphological information. \cite{bhingardive-etal-2014-semi} proposed a semi-automatic approach to enhance this Sanskrit wordnet, using mapping from an existing bilingual Sanskrit English dictionary and the Princeton WordNet to populate the Sanskrit synsets.

\begin{table}[h]
    \caption{NLP Research Works using Rule-based strategy for Indic Languages}
    \label{tab:rule-based}
    \centering
    \begin{tabular}{p{2.5cm}|p{1.5cm}|p{1cm}|p{1.5cm}|p{7cm}}
        \toprule
            & \multicolumn{2}{p{3cm}}{Focused Task} & \\
        \cmidrule{2-4}
        Model & Machine Translation & Parsing & Resource Creation & Approach\\
        \midrule
        \cite{bharati-sangal-1990-karaka} & $\checkmark$ & $\checkmark$ &  & Core parsing formulated as integer programming problem \\
        \cite{bharati1991local} &  & $\checkmark$ &  & Local word grouping (verb and noun groups) used as a pre-parsing strategy\\
        \cite{bharati1993parsing} & & $\checkmark$ & & Constraint based parsing using the Paninian framework\\
        \cite{ANGLABHARTI} & $\checkmark$ &  & & Uses a `pseudo-target' applicable to a group of Indic languages, later used for text generation \\
        \cite{rao1998natural} & $\checkmark$ & & & Structural and Lexical transfer, tagged with semantic knowledge\\
        \cite{dave2001interlingua} & $\checkmark$ & & & Interlingua-based translation using the Universal Networking Language (UNL) \\
        \cite{bharati2003anusaarakaovercominglanguagebarrier} & $\checkmark$ & & & Human-machine aided MT via a language-independent text image\\
        \cite{begum-etal-2008-dependency} & & $\checkmark$ & $\checkmark$ & Syntactico-semantic dependency annotation based on the Paninian framework\\
        \cite{gupta2008rule} & & $\checkmark$ & $\checkmark$ & Automatic annotation using syntactic cues, following the Paninian dependency framework. \\
        \cite{bharati2009simple} & & $\checkmark$ & & Corpus-based parsing to identify inter and intra-chunk dependency relations\\
        \cite{bharati2008two} & & $\checkmark$ & & Data-driven dependency parsing enhanced with two features: conjoined vibhakti-label feature and minimal semantics\\
        \cite{Husain2009data} & & $\checkmark$ & & Modular cascaded data-driven dependency parsing for Hindi\\
        \cite{bharati2009constraint} & & $\checkmark$ & & Two stage constraint-based dependency parsing to identify inter and intra-clausal dependencies\\
        \cite{bharati2003leril} & & & $\checkmark$ & Collaborative development of lexical resources using crowd sourcing and open source tools \\
        \cite{kulkarni2010introducing} & & & $\checkmark$ &  Construction of Sanskrit wordnet using the expansion approach, based on Hindi wordnet\\
        \cite{bhingardive-etal-2014-semi} & & & $\checkmark$ & Sanskrit wordnet extension using Princeton wordnet and Hindi wordnet \\
        \bottomrule
    \end{tabular}
\end{table}

\subsection{Corpus-based Indic NLP}
Corpus-based or statistical NLP relies on the analysis of large, structured corpora to automatically derive linguistic patterns or statistical probabilities. Corpus-based NLP uses empirical data to perform tasks such as tagging, parsing, and translation. Unlike the standard practice that assumes a fixed sentence structure, statistical NLP was used for morphological analysis of the morphologically rich and free word order Indic languages. Apart from this, research focused on hybrid parsing for Indic languages by integrating rule-based patterns with statistical models to overcome the drawbacks of the traditional constituency parsing technique.

In the case of the Statistical Machine Translation (SMT), three approaches were used: word-based, phrase-based, and hierarchical phrase-based translation. While the input sentence was translated word by word and then arranged to get the target sentence in the first approach, each source and target sentence was divided into different phrases and aligned using patterns in the phrase-based approach. The hierarchical phrase-based translation approach used hierarchical phrases with recursive structures instead of simple phrases. A few studies also combined both rule-based translation and corpus-based translation approaches to achieve better results. In Example-based translation, the target sentence was formed from the source sentence using pre-translated examples. This used parallel corpora, which contained sentence pairs with a source sentence and its translation in the required language.

\textit{Shata-Anuvadak} by~\cite{kunchukuttan-etal-2014-shata} was the largest effort at the time and included a collection of phrase-based Statistical Machine Translation (SMT) systems for 110 language pairs. The study used the Indian Language Corpora Initiative (ILCI) parallel corpus of 11 Indian languages, containing roughly 50000 parallel sentences belonging to the health and tourism domains. An objective of the study was to understand the relation between the accuracy of translation and the language family involved. It also aimed to explore whether the shared characteristics of Indian languages can reduce the efforts and resources required for building technology. The provided results called for customized approaches for language family pairs, as the results of translation were better in the case of a few Indic languages, where a higher level of accuracy was seen compared to a few other Indic languages. Similar results were seen in experiments using increased corpus size, and while the translation performance was better in the case of a few languages, the improvement was minimal for other languages with rich morphology. The reordering on the source side helped languages such as Tamil, Telugu, Kannada, and Malayalam a little better than languages like Hindi, whereas the post-editing with transliteration improved the translation quality significantly for the language pairs with scripts derived from the \textit{Brahmi} script. 

Incorporation of syntactic and morphological information has been shown to result in significant improvements in phrase-based SMT. For English-Hindi translation, reordering the English source sentence according to Hindi syntax and using a simple suffix separation program for Hindi can be effective~\cite{ramanathan2008simple}. In English to Hindi translation, the case markers and suffixes in Hindi are predominantly determined by the combination of suffixes and semantic relations on the English side. Hence, the challenges posed by features of Hindi, such as free-word order and rich morphology for translation, are shown to be alleviated by augmenting the aligned corpus of English-Hindi with the correspondence of English suffixes and semantic relations with Hindi suffixes and case markers~\cite{ramanathan-etal-2009-case}. Still, translation between language pairs with huge syntactic differences requires re-ordering, which in turn needs an understanding of the word order in the respective language. For English-Hindi translation, research by \cite{ramanathan2011clause} demonstrated that the translation quality can be significantly improved by performing clause-wise translation. \cite{mtil17} reported that using rules to transfer the structure of the source sentences before training and translation can lead to a better translation quality. Additionally, they showed that suffix separation can be used to tackle the morphological divergence between English and highly agglutinative Indian languages.

Along with the model development efforts, various semantic resources were also created. A few of them are IndoWordNet~\cite{bhattacharyya-2010-indowordnet}, Bodo Wordnet~\cite{sarma2010wordnet}, Tamil Wordnet~\cite{rajendran2002tamil}, Kannada Wordnet~\cite{sahoo2003kannada}, Sanskrit Wordnet~\cite{kulkarni2010introducing}, Assamese Wordnet~\cite{sarma2010foundation}, and Punjabi Wordnet~\cite{narang2013development}. Other resources, such as n-grams, were also proposed for Indic languages~\cite{Majumder2002NgramAL}.

Multiple approaches for Part-of-Speech tagging were also proposed using techniques such as Hidden Markov Model (HMM)~\cite{bengalihybridpos, tamilpos} and Maximum Entropy Markov Model (MEMM)~\cite{hindiposmemm, dalal2007building}. Table~\ref{tab:corpus-based} lists the corpus-based research works in IndicNLP.

\begin{table}[]
    \caption{Research works using the corpus-based NLP techniques for Indic languages}
    \label{tab:corpus-based}
    \centering
    \begin{adjustbox}{width=\textwidth}
    \begin{tabular}{p{.2\linewidth}|p{.2\linewidth}|p{.6\linewidth}}
        \toprule
        \textbf{Paper} & \textbf{Language(s)} & \textbf{Approach} \\
        \midrule
        \multicolumn{3}{c}{Parsing} \\
        \midrule
        \cite{bengalihybridpos} & Bengali &  Hidden Markov Model \\
        \cite{singh2005hmm} & Hindi & Hidden Markov Model \\
        \cite{hindiposmemm} & Hindi & Maximum Entropy Markov Model \\
        \cite{dalal2007building} & Hindi &  Maximum Entropy Markov Model \\
        \cite{Gorla2008Dependency} & Hindi & Normalized Conditional Mutual Information (NCMI) \\ 
        \cite{tamilpos} & Tamil & Hidden Markov Model \\
        \cite{bharati2009constraint} & Hindi & Constraint based hybrid parsing   \\
        \cite{bharati2009two} & Hindi & Constraint based Two-stage parsing \\
        \cite{5329357} & Tamil & Conditional Random Fields \\
        \cite{kumar2010Sanskrit} & Sanskrit & Automatic segmentation of compounds \\
        \cite{ramasamy2011tamil} & Tamil & Dependency parsing \\
        \midrule
        \multicolumn{3}{c}{Machine Translation (MT) or Transliteration} \\
        \midrule
        \cite{1182307} & Bengali, Assamese & Example-based MT \\   
        \cite{ramanathan2008simple} & English, Hindi & Using syntactic and morphological information \\
        \cite{ramanathan-etal-2009-case} & English, Hindi & Using case markers and inflections \\
        \cite{Chaudhury2010Translation} & English, Hindi & Shallow parsing, deep parsing\\
        \cite{ahsan2010coupling} & English, Hindi & Coupling rule-based and statistical MT\\
        \cite{visweswariah2011word} & English, Hindi, Urdu  & Source side sentence reordering \\
        \cite{ramanathan2011clause} & English, Hindi & Clause-based translation with reordering constraints \\
        \cite{kunchukuttan2012partially} & Multilingual & Source side phrase reordering \\
        \cite{kriya} & English, Hindi & Hierarchical phrase-based MT \\
        \cite{kunchukuttan-etal-2014-shata} & Multilingual & Phrase-based statistical MT \\
        \cite{kunchukuttan2015brahmi} & Multilingual & Parallel transliteration corpora \\
        \cite{mtil17} & Multilingual & Using pre-ordering and suffix separation \\
        \cite{tholpadi2017corpus} & Multilingual & Comparable corpora-based translation correspondence induction \\
        \midrule
        \multicolumn{3}{c}{Semantic Processing} \\
        \midrule
        \cite{reddy2009words} & Hindi & Semantic category labeling \\
        \cite{statisticalma} & Multilingual & Statistical morphology analysis \\
        \bottomrule
    \end{tabular}
    \end{adjustbox}
\end{table}

\subsection{Indic NLP using Deep Learning}
The emergence of Deep Learning (DL) revolutionized Indic NLP by moving the focus from manual feature engineering that was common in the statistical NLP era to automatic representation learning. Characterized by the use of embeddings and attention mechanisms to capture the semantic details that were missed by the previous models, the usage of deep learning models for Indic languages was driven by the need to handle complex morphology, script dissimilarity, and the deficiency of huge parallel corpora for regional language pairs. Also, the dense vector representation used by these models addressed the issue of the formation of the Out-of-Vocabulary (OOV) words in Indic languages, often due to inaccurate sub-word level representation. DL was used for tasks such as Neural Machine Translation (NMT), generating Multilingual representations, and language generation. Summary of the Indic NLP works using deep learning is provided in Table~\ref{tab:deeplearning}.

\subsubsection{Embedding Representations and Indic Language Models}
Pre-trained embeddings (monolingual and cross-lingual) for 14 Indic languages were generated by~\cite{kumar-etal-2020-passage} using both contextual and non-contextual approaches. IndicFT~\cite{kakwani-etal-2020-indicnlpsuite} developed a set of pre-trained word embeddings for 11 Indic languages trained on the massive IndicCorp dataset~\cite{kakwani-etal-2020-indicnlpsuite}. The core of IndicFT's strategy relies on the FastText model, which utilizes character n-grams (subword information) rather than treating whole words as indivisible atomic units. This method was explicitly chosen because Indian languages are morphologically rich and agglutinative, meaning words often change form with complex suffixes. By breaking words down into smaller character chunks (n-grams) during training, IndicFT could generate embeddings for rare or unseen words by summing the representations of their sub-parts, allowing the model to effectively capture semantic similarities between words that share common roots.

To address the need for robust Natural Language Understanding (NLU) across Indic languages, the IndicBERT~\cite{kakwani-etal-2020-indicnlpsuite} model was developed using the compact ALBERT architecture trained on the 8.9 billion token IndicCorp dataset. Distinct from the character n-gram approach used in IndicFT, IndicBERT employed a SentencePiece~\cite{kudo2018sentencepiece} tokenizer with a large shared vocabulary of 200,000 tokens, specifically designed to accommodate the diverse scripts and wide lexical variety of the 12 supported languages (11 Indian languages plus English). Despite having significantly fewer parameters than massive multilingual baselines like multilingual BERT~\cite{devlin2019bertpretrainingdeepbidirectional} (mBERT) and XLM-R~\cite{DBLP:journals/corr/abs-1911-02116} (12M vs. 110M+), IndicBERT demonstrated superior performance on several IndicGLUE~\cite{kakwani-etal-2020-indicnlpsuite} benchmark tasks.

Complementing model development, \cite{karthika2025multilingual} presented a comprehensive intrinsic evaluation of tokenization specifically for 17 Indian languages. This work contrasted the efficacy of Byte Pair Encoding (BPE) and Unigram Language Model (ULM), finding that ULM provides superior morphological alignment, adhering more closely to the linguistic segmentation required for morphologically rich and agglutinative Indic languages. The study further highlighted the critical impact of pre-processing; tokenizers trained on normalized corpora (standardizing Unicode and script-specific characters) consistently yielded lower fertility scores, indicating significantly more efficient segmentation. Additionally, they demonstrated that cluster-based training, grouping languages by typological similarity, significantly reduced the Word Fragmentation Rate (WFR) for underrepresented languages compared to joint training strategies, thereby mitigating the dominance of high-resource languages in the shared vocabulary.

The design of multilingual vocabularies presents significant trade-offs. While IndicBERT adopted a 200k vocabulary to ensure broad coverage, \cite{karthika2025multilingual} observed that increasing vocabulary size from 32k to 256k consistently improves intrinsic metrics such as fertility and fragmentation rate. However, this work identified a diminishing return in cross-lingual alignment: vocabulary overlap between languages increased up to 128k but decreased at 256k, suggesting that larger vocabularies may begin to include arbitrary, non-shared tokens. Furthermore, the study revealed promising zero-shot transfer capabilities; multilingual tokenizers trained on high-resource languages were able to effectively segment unseen, extremely low-resource dialects like Awadhi, Bhojpuri, and Magahi, achieving fertility scores (e.g., 1.3–1.4) comparable to those of high-resource languages.

To address the limitations of massive multilingual models in representing Indian languages, \cite{Khanuja2021MuRILMR} proposed Multilingual Representations for Indian Languages (MuRIL), which employs a cased WordPiece vocabulary of 197,285 tokens generated specifically from upsampled Indian language corpora. Unlike standard multilingual models that often lowercase input, MuRIL preserved case to retain accent information, which is semantically significant in many Indic languages. The vocabulary generation process explicitly utilized upsampling to ensure that low-resource languages are adequately represented, preventing high-resource languages from dominating the subword inventory.

Empirical analysis demonstrated that MuRIL achieves a significantly lower fertility ratio (average subwords per word) across all 17 supported languages compared to mBERT. High fertility ratios, common in mBERT due to its vocabulary being approximately 78\% Latin script, often lead to the over-fragmentation of Indic words into meaningless characters, thereby degrading semantic understanding. In contrast, MuRIL allocated a substantially higher proportion of its vocabulary to Indic scripts (e.g., Devanagari, Bengali, Tamil), ensuring that native words are tokenized into meaningful subword units rather than arbitrary character sequences. Furthermore, MuRIL addressed the prevalence of code-mixing and transliteration in the Indian digital landscape by explicitly including transliterated data (native languages written in Latin script) during pre-training. While mBERT's vocabulary didn't provide sufficient coverage for transliterated forms, MuRIL's tokenizer could effectively recognize and segment transliterated terms, contributing to its superior performance on tasks involving informal text and code-switching.

Recent experimental evidence underscores the significant performance advantage of language-specific pre-training over cross-lingual transfer learning for complex morphological tasks. \cite{dasari2023transformer} provided a comprehensive evaluation of Telugu morphological analysis and demonstrated that a monolingual Transformer model (\textit{BERT-Te}), trained from scratch on a dedicated corpus of roughly 8 million Telugu sentences, consistently outperformed massive multilingual baselines including mBERT, XLM-R, and IndicBERT. This finding challenges the prevailing assumption that multilingual models are sufficient for low-resource languages, particularly those with agglutinative morphology.

The performance gap is most evident in the extraction of fine-grained grammatical features. The study reported that the monolingual \textit{BERT-Te} achieved an F1 score of 0.778 for Gender tagging, significantly surpassing IndicBERT (0.527) and XLM-R (0.624). Similarly, for the category of Person, the monolingual model achieved an F1 score of 0.704, whereas IndicBERT struggled with a score of 0.475. The authors attributed this disparity to ``concentrated language instruction''; while multilingual models dilute their capacity across hundreds of languages, the monolingual architecture captures the intricate inflectional nuances and structural complexities specific to the target language. Consequently, for tasks requiring deep morphological understanding, domain-specific training provides a quantifiable advantage over generalized multilingual representations.

\subsubsection{Sequence Modeling Tasks}

Recurrent Neural Networks (RNNs), Long Short-Term Memory (LSTM) networks, and Transformer architectures were widely used for handling the sequential nature of text. These models performed better at `remembering' long-term dependencies in a sentence compared to the previous modeling approaches. This capability made them better suited for free-word order Indic languages. LSTMs were widely used for POS tagging and Named Entity Recognition (NER) tasks.

POS tagging leveraged multilingual Transformer models to overcome data scarcity. Addressing the lack of resources for extremely low-resource languages, \cite{kumar-etal-2024-part} presented the first Universal Dependencies (UD)-compliant POS tagging datasets for Angika and Magahi, alongside a new parallel dataset for Bhojpuri. This work highlighted the critical challenge of sub-optimal tokenization in multilingual models like MuRIL and XLM-R, where over-fragmentation of words in low-resource languages leads to poor tagging performance. To mitigate this, this work proposed novel ``Look-back'' and ``Look-back-with-score'' techniques, which utilized the tag of the first sub-word token or the token with the maximum logit score to represent the entire word. Experiments demonstrated that Indic-specific models such as MuRIL outperform massive multilingual models in zero-shot settings due to reduced language interference, and that cross-lingual transfer from Hindi is highly effective when alignment errors are minimized.

Complementing the development of taggers, recent interpretability studies have investigated how transformer models encode POS and morphological properties. \cite{aravapalli2024indicsenteval} introduced \textit{INDICSENTEVAL}, a benchmark for probing linguistic properties across six Indic languages. Unlike standard tagging tasks, this work utilized POS information (such as subject/object number and verb attributes) as diagnostic probing tasks to assess the internal representations of nine multilingual models. The findings from this study reinforce the superiority of Indic-specific models (e.g., MuRIL, IndicBERT) in capturing semantic POS properties compared to universal models. However, a perturbation analysis revealed a counter-intuitive insight: universal models (e.g., InfoXLM~\cite{chi-etal-2021-infoxlm}, mT5~\cite{DBLP:journals/corr/abs-2010-11934}) often exhibit greater robustness to input noise, such as the dropping of nouns and verbs, suggesting that while Indic-specific models are more accurate for clean text, they may rely more heavily on specific lexical cues and word order.

A significant innovation in recent Transformer-based architectures is the move away from explicit, rule-based stemming towards ``implicit stemming" mechanisms embedded within subword tokenization. \cite{kakwani-etal-2020-indicnlpsuite} demonstrated this with \textit{IndicBERT}, a multilingual model pre-trained on large-scale corpora spanning 11 Indian languages. Unlike traditional pipelines that require a dedicated pre-processing step to strip suffixes using fixed linguistic rules, IndicBERT utilized the SentencePiece tokenizer to decompose words into statistically frequent subword units (e.g., splitting a fused word into root and affix components). This approach allowed the model to learn morphological representations and root-affix relationships directly from data, effectively bypassing the need for handcrafted stemmers while handling Out-of-Vocabulary (OOV) terms through subword composition.

\begin{table}[]
    \caption{Indic NLP Works using Deep Learning}
    \label{tab:deeplearning}
    \centering
    \begin{tabular}{p{0.15\linewidth}|p{0.35\linewidth}|p{0.5\linewidth}}
    \toprule
     Paper / Initiative  & Core Approach & Unique Features  \\
     \midrule
     \cite{kumar-etal-2020-passage} & Generated both contextual and non-contextual pre-trained embeddings & \begin{itemize}[leftmargin=*]
     \item Covered 14 Indic languages 
     \end{itemize}\\
     IndicFT \cite{kakwani-etal-2020-indicnlpsuite} & Utilized the FastText model which relies on character n-grams (subword information) rather than whole words & \begin{itemize}[leftmargin=*]
         \item Covered 11 Indic languages trained on the IndicCorp dataset
         \item Explicitly handles morphologically rich and agglutinative structures by generating embeddings for rare or unseen words via sub-parts
     \end{itemize} \\
     IndicBERT \cite{kakwani-etal-2020-indicnlpsuite} & Employed a compact ALBERT architecture using a SentencePiece tokenizer with a large shared vocabulary of 200,000 tokens & \begin{itemize}[leftmargin=*]
         \item Supported 12 languages (11 Indian + English)
         \item Outperformed massive baselines (mBERT, XLM-R) on IndicGLUE despite having significantly fewer parameters (12M vs. 110M+)
     \end{itemize} \\
     \cite{karthika2025multilingual} & Provided an intrinsic evaluation of tokenization, contrasting Byte Pair Encoding (BPE) and Unigram Language Model (ULM) & \begin{itemize}[leftmargin=*]
         \item Evaluated 17 Indian languages
         \item Demonstrated zero-shot transfer capabilities to segment unseen dialects (Awadhi, Bhojpuri, Magahi)
     \end{itemize} \\
     MuRIL \cite{Khanuja2021MuRILMR} & Employs a cased WordPiece vocabulary of 197,285 tokens generated from upsampled Indian language corpora & \begin{itemize}[leftmargin=*]
         \item Preserved text case to retain semantically significant accent information
         \item Explicitly included transliterated data during pre-training to handle code-mixing and code-switching
         \item Achieved a lower fertility ratio compared to mBERT by allocating more space to native Indic scripts
     \end{itemize} \\
     BERT-Te \cite{dasari2023transformer} & Trained a language-specific, monolingual Transformer model from scratch on roughly 8 million Telugu sentences & \begin{itemize}[leftmargin=*]
         \item Consistently outperformed massive multilingual baselines on Telugu morphological analysis
         \item Achieved significantly higher F1 scores for fine-grained grammatical features like Gender and Person tagging via "concentrated language instruction"
     \end{itemize} \\
     INDICSENTEVAL \cite{aravapalli2024indicsenteval} & Introduced a diagnostic probing benchmark using POS information to assess the internal representations of 9 multilingual models & \begin{itemize}[leftmargin=*]
         \item Evaluated 6 Indic languages
         \item Reinforced the superiority of Indic-specific models in capturing semantic POS properties, while revealing that universal models (e.g., InfoXLM, mT5) showed greater robustness to input noise
     \end{itemize} \\
     \bottomrule
    \end{tabular}
\end{table}

\subsection{Foundation Models}

The development of Natural Language Understanding (NLU) for Indic languages has advanced through three distinct phases: initial resource creation, architectural specialization for linguistic nuances, and rigorous, human-centric evaluation. This progression is evident across key NLU tasks such as Named Entity Recognition (NER), Question Answering (QA), Sentence Retrieval, Natural Language Inference (NLI), and Paraphrase Detection.

\cite{kakwani-etal-2020-indicnlpsuite} established the first comprehensive baseline with the \textsc{IndicNLP Suite}. This work introduced \textsc{IndicBERT} v1 (an ALBERT-based model) and the \textsc{IndicGLUE} benchmark. For NER, the model was evaluated on the silver-standard \textsc{WikiAnn} dataset, achieving an F1 score of 64.47, but trailing the mBERT baseline of 73.24 due to limited model capacity. In Question Answering, the suite utilized a cloze-style multiple-choice format, where the model achieved an accuracy of 41.91, demonstrating the feasibility of knowledge extraction in Indic languages. The work also benchmarked cross-lingual sentence retrieval using the \textsc{CVIT-Mann Ki Baat} dataset (Precision@10: 27.12) and introduced coreference-focused inference via the \textsc{Winograd NLI} task (Accuracy: 56.34 on Hindi). Paraphrase detection was evaluated on the \textsc{Amritha} dataset, with the model scoring 93.11 accuracy on Hindi, setting a strong initial precedent for classification tasks.

Addressing the limitations of general multilingual models, \cite{Khanuja2021MuRILMR} introduced \textsc{MuRIL}, a BERT-based model augmented with translation and transliteration data to handle code-mixing and script diversity. This specialization led to state-of-the-art results on the \textsc{XTREME} benchmark. In NER, \textsc{MuRIL} achieved an F1 score of 77.2, significantly outperforming global baselines like mBERT and XLM-R on structural tasks. For Question Answering, it demonstrated superior reading comprehension on extractive datasets like \textsc{XQuAD} (Hindi F1: 73.9) and \textsc{TyDiQA} (Avg F1: 75.4). The model also excelled in cross-lingual tasks, achieving high accuracy in Natural Language Inference (IndicXNLI: 74.1) and Paraphrase Detection (72.4), leveraging its transliteration-aware pretraining to bridge semantic gaps across languages.

The most recent phase, led by \cite{doddapaneni2023towards}, prioritized evaluation integrity with the \textsc{IndicXTREME} benchmark. This work introduced \textsc{IndicBERT} v2, trained on the massive \textsc{IndicCorp} v2. A key shift was the use of human-verified datasets like \textsc{Naamapadam} for NER, where \textsc{IndicBERT} v2 scored 72.4, beating global baselines but trailing \textsc{MuRIL}. In Question Answering, the model matched the strong \textsc{MuRIL} baseline on the manually curated \textsc{IndicQA} dataset (F1: 47.7) while vastly outperforming XLM-R. \textsc{IndicBERT} v2 showed its strongest gains in Sentence Retrieval, scoring 69.4 on the high-quality \textsc{FLORES} dataset, a 14.5-point improvement over prior models. However, in "strict" semantic tasks like Paraphrase Detection on the new \textsc{IndicXParaphrase} dataset, it scored 56.9, again highlighting that while larger corpora improve general alignment, specific pretraining strategies remain crucial for resolving fine-grained semantic nuances.

Bharat Parallel Corpus Collection (BPCC), the largest publicly available parallel corpus for Indic languages, was created by~\cite{gala2023indictrans2highqualityaccessiblemachine}. In addition to this, the work also introduced an n-way parallel benchmark covering 22 Indian languages, featuring diverse domains, Indian-origin content, and source-original test sets. \textit{IndicTrans2}, the first model to support all 22 languages, was also created as a part of this work. The lack of parallel training data, robust benchmarks, and translation models spanning all 22 languages was overcome with this work.

\subsubsection{Indic NLP using Generative AI}
% Explain in detail the generative models for Indic languages
Recently, several groups have been leading efforts to build generative LLMs that cater to the diverse linguistic landscape of India. A few notable initiatives are Sarvam AI\footnote{\hyperlink{https://www.sarvam.ai/}{https://www.sarvam.ai/}}, BharatGen\footnote{\hyperlink{https://bharatgen.com/}{https://bharatgen.com/}}, and AI4Bharat\footnote{\hyperlink{https://ai4bharat.iitm.ac.in/}{https://ai4bharat.iitm.ac.in/}} among others.

BharatGen has introduced pioneering works, including \cite{docbodh} - a GenAI suite for Indic Document understanding, \cite{Param} - a text model trained on India-centric data, and can understand and generate human-like text in multiple Indian languages and dialects. Sarvam AI has released powerful models, including \cite{Sarvam-M} - a multilingual, hybrid-reasoning, text-only model built on Mistral-Small\footnote{\hyperlink{https://mistral.ai/news/mistral-small-3-1}{https://mistral.ai/news/mistral-small-3-1}}, following Sarvam-1 - a language model with 2-Billion parameters specifically optimized for Indian languages. Other noteworthy models are \cite{Sarvam-Translate} - providing comprehensive coverage of all scheduled Indian languages with formal translation style, and \cite{Mayura} - designed to convert text between English and Indian languages while preserving meaning and context.

In addition to this, initiatives such as Bhashini\footnote{\hyperlink{https://bhashini.gov.in/}{https://bhashini.gov.in/}} are launched by the government with an aim to bridge the language gap and enable access to digital services and content in regional languages. Bhashini provides translation services across more than 36 languages and also hosts a unified hub\footnote{\hyperlink{https://bhashini.gov.in/vatika}{https://bhashini.gov.in/vatika}} meant for discovering, exploring, and contributing AI models and datasets tailored for India's diverse linguistic - technological scenario.

The development of foundation and instruction-tuned models for Indic languages has accelerated since late 2023. Some of them are Gajendra~\cite{gajendra}, Airavata~\cite{gala2024airavataintroducinghindiinstructiontuned}, Krutrim LLM~\cite{kallappa2025krutrimllmmultilingualfoundational}, Tamil Llama~\cite{balachandran2023tamilllamanewtamillanguage}, Bharat GPT~\cite{bharatgpt}, and Navarasa~\cite{navarasa}. Table~\ref{tab:foundation} lists the generative AI works using Indic languages.

The training datasets for prominent Indic language models and initiatives often come from a combination of large-scale web scraping, synthetic data generation~\cite{manoj2025bhashakritikabuildingsyntheticpretraining}, curated translation pairs~\cite{karthika2026samasamayikparalleldatasethindisanskrit}, and crowdsourced speech~\cite{javed-etal-2024-indicvoices, Bhashadaan}. To the best of our knowledge, the representation of knowledge from indigenous communities is little when compared to that from the conventional sources. Directions towards bridging this gap are discussed in Section~\ref{sec:language}.

\begin{table}[]
    \caption{Indic NLP Foundation Models}
    \label{tab:foundation}
    \centering
    \begin{tabular}{p{0.15\linewidth}|p{0.2\linewidth}|p{0.6\linewidth}}
    \toprule
    Model / Initiative  & Core Approach & Key Results  \\
    \midrule
    IndicNLP Suite \cite{kakwani-etal-2020-indicnlpsuite} & Established the first comprehensive NLU baseline using IndicBERT v1 and the IndicGLUE benchmark & \begin{itemize}[leftmargin=*]
        \item Evaluated NER on the silver-standard WIKIANN dataset (F1: 64.47)
        \item Benchmarked tasks like Cloze-style Question Answering, cross-lingual sentence retrieval (CVIT-MANN KI BAAT), and Winograd NLI
    \end{itemize}\\
     MURIL \cite{Khanuja2021MuRILMR} & Specialized BERT-based model augmented with translation and transliteration data & \begin{itemize}[leftmargin=*]
         \item Achieved state-of-the-art results on the XTREME benchmark
         \item Outperformed global baselines in structural tasks like NER (F1: 77.2) and extractive QA (XQUAD Hindi F1: 73.9; TyDi-QA Avg F1: 75.4)
     \end{itemize} \\   
     % IndicXTREME \cite{doddapaneni2023towards} & Prioritized evaluation integrity using the IndicXTREME benchmark and IndicBERT v2 (trained on IndicCORP v2) & \begin{itemize}[leftmargin=*]
     %     \item Shifted focus toward human-verified datasets like Naamapadam~\cite{mhaske-etal-2023-naamapadam} for NER and IndicQA~\cite{singh2025indicqabenchmarkmultilingual} for QA
     %     \item Achieved a 14.5-point improvement in Sentence Retrieval on the high-quality FLORES dataset
     % \end{itemize}\\
     BPCC \& IndicTrans2 \cite{gala2023indictrans2highqualityaccessiblemachine} & Created the Bharat Parallel Corpus Collection (BPCC)—the largest public parallel corpus for Indic languages—and developed the IndicTrans2 mode & \begin{itemize}[leftmargin=*]
         \item  Introduced an n-way parallel benchmark featuring diverse domains and Indian-origin content
         \item IndicTrans2 became the first translation model to natively support all 22 scheduled Indian languages
     \end{itemize} \\
    BharatGen & Introduced a suite of specialized India-centric generative models & \begin{itemize}[leftmargin=*]
        \item DocBodh~\cite{docbodh}: A GenAI suite for Indic document understanding
        \item Param~\cite{Param}: A text model trained on India-centric data capable of understanding and generating regional text and dialects
    \end{itemize} \\
    Sarvam & Released a series of foundational text and specialized language models & \begin{itemize}[leftmargin=*]
        \item Sarvam-1~\cite{Sarvam_1}: A 2-Billion parameter language model optimized for Indian languages
        \item Sarvam-M~\cite{Sarvam-M}: A multilingual, hybrid-reasoning text model built on Mistral-Small
        \item Sarvam Translate~\cite{Sarvam-Translate}: Open-weights model for formal translation across 22 scheduled languages
        \item Mayura~\cite{Mayura}: Designed to convert text between English and Indian languages while preserving context
    \end{itemize} \\
    Bhashini & Government-led initiative serving as a unified digital ecosystem hub & \begin{itemize}[leftmargin=*] 
    \item Provides translation services across more than 36 languages to bridge the digital regional language gap
    \end{itemize} \\
    % Instruction-Tuned / LLM Initiatives (Late 2023–2025) & Development of foundational instruction-tuned LLMs targeting specific or multilingual Indic landscapes & \begin{itemize}[leftmargin=*] 
    % \item Includes Gajendra AI (Hindi-Hinglish-English model), Airavata (Hindi instruction-tuned), Krutrim LLM, Tamil Llama, Bharat GPT, and Navarasa (Gemma fine-tuned collection) 
    % \end{itemize} \\
    \bottomrule
    \end{tabular}
\end{table}

\subsection{Speech Recognition for Indic Languages}

\begin{table}[h]
    \caption{Key Projects and Models in Indic ASR}
    \label{tab:indic_asr_projects}
    \centering
    \begin{tabular}{l|p{10cm}}
    \toprule
    \textbf{Project / Model Name} & \textbf{Description} \\ 
    \midrule
    IndicWav2Vec \cite{javed2021buildingasrsystemsbillion} & Using Wav2Vec-like model architecture along with unlabeled data to pre-train the model. Achieved state-of-the-art results on 9 Indic languages, including on very low-resource Indic languages such as Sinhala and Nepali \\
    Vakyansh \cite{chadha2022vakyanshasrtoolkitlow} & An initiative focused on building an end-to-end toolkit for ASR using speech data in 23 Indic languages. Contributed ASR models in 18 Indic Languages, along with post-processing utilities such as Punctuation restoration and Inverse Text Normalization \\ 
    Shrutilipi \cite{bhogale2022effectivenessminingaudiotext} & A large-scale dataset mined from public archives (e.g., All India Radio), containing over 6,400 hours of labeled audio across 12 Indian languages\\ 
    IndicSUPERB \cite{javed2022indicsuperbspeechprocessinguniversal} & Kathbath, a large scale ASR dataset for 12 Indian languages used to create a benchmark for ASR, language identification, speaker identification, and speaker verification across the Indic language family  \\
    VAANI \cite{pulikodan2026vaanicapturinglanguagelandscape} & A dataset designed to capture diverse linguistic variations, accents, and code-switching, essential for dialectal ASR evaluation. Contains 2,043 hours of transcribed audio spanning 105 languages from 28 states and 3 union territories, many of these languages are being represented at this scale for the first time  \\ 
    Saaras (V3) \cite{Saaras_V3} & Streaming speech recognition supporting 22 Indian languages, featuring low-latency decoding and support for code-mixed speech \\
    \bottomrule
    \end{tabular}
\end{table}

Research in Automatic Speech Recognition (ASR) for Indic Languages has achieved significant progress, paralleling the works in NLP with large foundational models like Whisper~\cite{radford2022robustspeechrecognitionlargescale}. ASR faces similar challenges to the ones explained in Section~\ref{sec:challenges} in addition to code mixing~\cite{8554413}, multilingual speech, and noisy recording. Table~\ref{tab:indic_asr_projects} lists the recent works in ASR for Indic languages in brief.

A critical bottleneck identified in recent studies for Indic ASR is the tokenization process, which disproportionately impacts low-resource languages compared to others. While high-resource languages benefit from extensive token sets in pre-trained models, low-resource languages suffer from limited vocabulary coverage. This deficiency forces the tokenizer to fragment words into numerous small subwords or characters, leading to increased sequence lengths and substantially slower auto-regressive inference speeds. To address these inefficiencies, \cite{tripathi2025enhancing} proposed a novel tokenization strategy that augments the existing Whisper vocabulary with language-specific Byte Pair Encoding (BPE) tokens. These tokens were derived from datasets representative of common Indian language sequences to reduce token fertility. The pre-trained model architecture was modified by expanding the dimension of the final token head layer, initializing new random weights for the added tokens while freezing the original weights to preserve multilingual capabilities.

Through ablation studies, an optimal threshold of 250 additional tokens per language was identified. This targeted vocabulary expansion significantly compressed token sequences; for example, token counts for Hindi and Malayalam sentences were reduced by approximately 30\% and 60\%, respectively. Empirical evaluations demonstrated that this reduction not only accelerated inference speeds outperforming optimized baselines but also consistently improved the Word Error Rate (WER) across model variants.

While these Indic ASR models are increasingly capable of addressing cultural nuances, they are far from being robust. Substantial progress is still required to bridge the gap between the performance in mainstream languages and the real linguistic diversity. The focus of future research must be on curation of representative datasets and design of models that inherently capture the regional variations, rather than fixing them into rigid, monolingual frameworks. A low-cost and easy-to-implement approach for ASR using colloquial speech is discussed in Section~\ref{sec:indicasr}.

\section{Challenges for Indic NLP}\label{sec:challenges}
% summarize that SOTA is for mainstream linguistic Indic NLP. Diglossia represents a unique cultural perspective and necessitates understanding and inclusivity
While Indic NLP research has made significant progress in recent times, persistent challenges pose challenges for building representative and inclusive language models for Indic languages. These challenges exist in the case of a majority of Indic languages and also other related languages, and are not observed in languages like English, which are used to build mainstream NLP systems. In this section, we discuss these challenges in detail and also discuss relevant research works that focused on finding solutions using specific cues such as the structure of Indic languages, the difference between dialects, etc.

% \tikzset{
%         my node/.style={
%             draw,
%             minimum width=1cm,
%             rounded corners=3,
%         }
% }
\begin{figure}[h]
    \centering
    \begin{forest}
        for tree={%
            my node,
            l sep+=5pt,
            edge={gray, thick},
        }
        [Challenges for Indic NLP, fill=blue!20
        [Complex Morphology, fill=pink]
        [Grammar Rules, fill=red!50]
        [Diglossia, fill=yellow!50]
        [Lack of Resources, fill=violet!50]
        ]
    \end{forest}
    \caption{Main Challenges for Research in Indic NLP}
    \label{fig:challenges}
\end{figure}
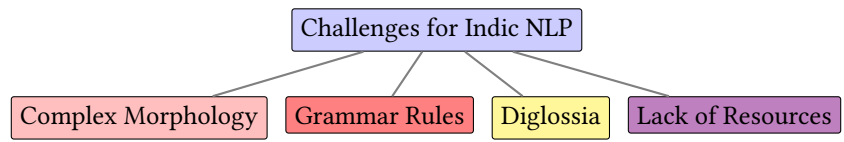

\subsection{Complex Morphology and Grammar Rules}
Some of the distinct features of morphology and grammar in the case of Indic languages, such as the \textit{kaaraka} rules along with the free word order, pose significant challenges when it comes to building language models for these languages. While there is a growing social and commercial need for the availability of language technologies in Indic languages, the scale and diversity of these languages make the design of inclusive language models difficult. Additionally, a few of the language features also need elaborate modeling. For example, compound verbs are verb-verb collocations where the polar verb adds the semantics and the vector verb acts as a modifier~\cite{slade2016compound}. Processing a sentence with such verbs requires a computational model to acquire both morphological and pragmatic understanding. Building language models that can understand and represent such fine nuances with precision is a complex task. \cite{bhattacharyya2019indic} discussed a list of challenges for Indic language computing and suggested some pointers for overcoming them. A few of the proposed solutions are a unified representation of Indic languages using a superset of sounds, interactive crowd sourcing for resource creation, and generic acoustic models and generic language models across various languages. In this section, we discuss the related studies that focus on using NLP techniques to model the linguistic features of Indic languages.

Indic languages are highly inflectional, and one root word can have multiple morphological variants. The agglutinative nature also makes morpheme boundaries hard to distinguish. Even though a set of rules can be used to identify the stem of a given word, this process requires linguistic expertise, is time-consuming, and language-specific. Researchers have explored the application of corpus-based techniques for morphological processing. \cite{kulkarninovel} proposed an improved indexing technique for identifying grammatically modified versions of the same root word. A fuzzy matching-based indexing mechanism inspired by prefix trees was used to index and retrieve different morphological forms of a term. This solution can be adapted for other similar languages with minimum modifications due to the similarities in their Unicode encoding and morphological behaviors. Another study by~\cite{DBLP:conf/wssanlp/Bhat12} found unsupervised morphological segmentation algorithms to perform well for the problem of morpheme boundary detection in Kannada. \cite{Ramanathan2003ALS} proposed a domain-independent and computationally inexpensive stemmer for the Hindi language. A suffix list was developed and used in this work. In the case of the Bengali language, \cite{majumder2007yass} explored the clustering-based approach, where equivalence classes of root words and their morphological variants were identified from a lexicon using string distance measures. Stemming performance using the approach provided results comparable to the standard rule-based stemmers and was observed to be effective for languages that are primarily suffixing. Unsupervised learning for acquisition of morphological knowledge from a text corpus was experimented with by~\cite{sharma2008acquisition} for the Assamese language. This approach was used to build a lexicon and subsequently used for morphological analysis. A formulation for derivational morphology, augmented Finite State Automata, was proposed along with a formalism for a lexicon writer for specifying a lexicon by~\cite{sengupta1996morphological}. This approach was also proposed to be useful as a building block for a spelling corrector in the case of the Bangla language. A finite state machine was used to develop a prototype morphological analyzer for the Kannada language~\cite{10.1007/978-3-540-77094-7_18}. 

Efficient morphological processing along with syntactic cues significantly improved the overall performance when incorporated into tasks such as SMT~\cite{ramanathan2008simple} and data-driven dependency parsing~\cite{ambati2010role, malladi2013statistical}. Another study found that augmenting the phrase table with all possible forms of a verb can improve the overall accuracy of a phrase-based MT system in the case of highly inflected languages~\cite{gandhe2011handling}. Morphological analysis has also been integrated with the task of tagging in multiple studies. Using a stemmer along with a morphological analyzer,~\cite{shrivastava2005morphology} performed Part-of-Speech (POS) tagging in the Hindi language. \cite{shrivastava2008hindi} showed that an HMM-based POS tagger was able to achieve reasonably good accuracy when stemming was performed as pre-processing. To overcome the lack of standardized tagsets,~\cite{sankaran2008designing} proposed a common POS tagset hierarchical framework for eight Indic languages.

\subsection{Diglossia}
Most of the Indic NLP applications focus on the formal language varieties. Along with this, the corpora used for training of Indic LLMs are dialect-neutral. Studies are being conducted to model the relation between the formal variety of a language and its dialects. \cite{mishra2010hindi} listed rules for phonology transfer from standard Hindi to a number of its prominent dialects, Bundeli, Bagheli, Kanauji and Awadhi. For Punjabi language and its dialects, Malwai and Doabi, a conversion system was proposed by~\cite{singh2015punjabi} using bilingual dictionaries and morphological conversion rules. In the case of the Tamil language, an approach for conversion from spoken dialectal language to standard written Tamil utilized Finite State Transducer (FST)~\cite{marimuthu2014automatic}. A framework for capturing synchronic variation by modeling the underlying diachronic variation was proposed by~\cite{choudhury2006rewrite}. This work used an ordered set of rewrite rules representing phonological changes that occurred during the evolution of dialects to derive current dialectal forms of the Bangla language from their classical counterparts.

Corpora collected in informal settings with unscripted, impromptu conversations can reveal the vibrant nature of languages and also reveal their characteristic patterns~\cite {dash2001we}. Human in the loop approach can be helpful in collating dialect-rich corpus from informal sources as shown by~\cite{10.1145/3632410.3632483}. This work utilized a small speech corpus collected from a rural community radio to build a colloquial text corpus. With the widespread availability of language models, there is a growing need to build dialect-rich corpora that can be used in turn to incorporate diverse knowledge into the language models.

\subsection{Lack of Resources}
% non-standard transliteration, non-standard storage
The lack of resources in Indic NLP refers to the multifaceted scarcity that spans data, computation, and linguistic expertise. While the Indian subcontinent is data-rich in terms of human interactions and knowledge systems, the digital footprint is disproportionately small compared to English. LLMs primarily rely on large amounts of high-quality data along with rich representations in the form of word embeddings. These word embeddings capture the contextual semantics of words and tokens, and are learned from huge text corpora with millions of words. However, the availability of such huge datasets in the case of Indic languages is limited, and this presents a challenge for the development of Indic language technologies. The largest publicly available Indic language datasets are of the order of billions of tokens\footnote{\href{https://huggingface.co/datasets?language=language:hi&sort=largest}{https://huggingface.co/datasets?language=language:hi\&sort=largest}}, whereas it is much higher for languages like English. Also, Indic languages have adopted multiple different encoding standards, such as ISCII~\cite{iscii}, Unicode~\cite{utf-8}, and font standards like ISFOC~\cite{isfoc}. These multiple standards have different software dependencies, and hence, uniform adaptation becomes difficult. In addition to this, there is no standard for transliteration among Indic languages and from Indic languages to other languages. Studies have proposed multiple approaches for Indic language transliteration~\cite{kunchukuttan2015brahmi, srivastava2013transliteration, diwakar2010transliteration, joshi2018indian, surana2008more}. However, there can still be variation in the representation of words. For example, the word for 'Thank you' in Hindi can be written as \textit{dhanyawad}, \textit{dhanywad}, \textit{dhanyavad}, or \textit{dhanyvad}. Large, labeled corpora are critically needed for Indic languages, and active research efforts are underway to address this gap.

Enabling Minority Language Engineering (EMILLE)~\cite{mcenery-etal-2000-emille}, Linguistic Data Consortium for Indian Languages (LDC-IL)~\cite{ldc-il} are a few of the initiatives aimed at creating a corpus in Indic languages. Studies have also experimented with various approaches to overcome the lack of data. For the task of translation,~\cite{philip2021revisiting} used iterative alignment to provide a large-scale sentence-aligned dataset using publicly available websites such as press releases by the government. For translation between Assamese and other Indic languages, a comparison between statistical and neural MT was made by~\cite{baruah2021low} to improve the translation performance in the case of the Assamese language. \cite{kumar2024part} contributed a Universal Dependency compliant parallel dataset for POS tagging in three Indic languages- Angika, Magahi, and Bhojpuri. Submissions to the low-resource Indic language translation task as part of the Conference on Machine Translation (WMT)~\cite{pakray2025findings} revealed important insights. A few of them are-

\begin{enumerate}
    \item a strong correlation exists between translation performance and the data size
    \item the effect of data quality and model optimization can outbalance the data quantity
    \item asymmetry in translation quality from one direction to the other
    \item models that perform well on one metric also perform well on other metrics of translation quality
\end{enumerate}

The shared linguistic features of Indic languages have also been useful in building language resources for low-resource languages. Studies have utilized resources of a related language with more resources to overcome the limitations for low-resource languages~\cite{singh2008estimating}. The similarity in scripts and sentence structure is shown to be effective in building language models for low-resource languages~\cite{khemchandani2021exploiting}, rather than direct training or using English as the pivot language. Even for translation, it is found that using a parent multilingual neural MT model and fine-tuning it on a low-resource language pair of interest performs better than standard neural MT~\cite{goyal2020efficient}.

% One of the reasons for the lack of good-quality corpora in Indic languages is their reliance on oral tradition. Knowledge is passed down through generations through verbal means such as storytelling, folklore, etc. Usage of such dialectal uncodified knowledge for building language technology requires linguistic expertise and manual effort for annotation. Very few research works have worked on building corpora from oral knowledge. \cite{10.1007/978-3-031-58502-9_1} used a human-in-the-loop approach for extracting a textual corpus from a small speech corpus collected in a rural setting with the help of a publicly available ASR model. Similar easy-to-use and economical approaches can be explored for collating corpora with knowledge in Indic languages.

Apart from the data quantity standpoint, the lack of specialized resources for Indic languages leads to technical inefficiencies. The issue of subword fragmentation, where a standard tokenizer breaks a single word into multiple tiny, meaningless fragments, is very much relevant in the case of Indic languages. This causes an increase in computation cost and latency. Hence, an English sentence with `n' tokens might have significantly more than fragments in the case of the Indic languages, also making it slower, more expensive, and meaningless. An approach was proposed by~\cite{brahma2025morphtokmorphologicallygroundedtokenization}, incorporating sandhi splitting to enhance the subword tokenization. It was shown that handling dependent vowels by forming a cohesive unit with other characters instead of occurring as a single unit leads to a reduction in fertility scores while maintaining performance in the language modeling task.

Techniques such as cross-lingual transfer~\cite{pawar-etal-2023-evaluating} and back-translation~\cite{das-etal-2025-investigating} have been adapted to mitigate the data scarcity issue. Crowd-sourcing initiatives like Bhashadaan\footnote{\href{https://bhashadaan.bhashini.co.in/bhashadaan/en/home}{https://bhashadaan.bhashini.co.in/bhashadaan/en/home}} are working on manually collecting data to train and to empower the linguistic diversity in Indic languages.

\section{Language and Culture}\label{sec:language}

Sections~\ref{sec:history} and \ref{sec:challenges} outlined the historical evolution and predominant challenges of Indic NLP, translating these insights into pluralistic AI requires new conceptual frameworks. To nudge the field to move forward, we propose \textit{Culture Sensing} - the process of understanding the worldviews and hermeneutics of a population to conduct meaningful communication with them, and also to enrich our own understanding of the world. Culture Sensing seeks to gather knowledge from various native discourses, regardless of the medium used, such as speech, text, and more. Unscripted, spontaneous data curation techniques are preferred to emphasize authenticity. The focus is on preserving the numerous credible hermeneutic schools and their associated methods of inquiry, as well as enhancing foundation models accordingly. In this section, we discuss the relation between language and culture, and how the inherent worldview in foundation models reflects the underlying culture. We also provide a demonstration of the Culture Sensing approach reflected in two different applications.

Language and Culture have a synergistic relation. Language acts as the primary vehicle through which cultural values, histories, and social systems are conserved and communicated. This relationship is complex in the context of the Indian subcontinent due to the immense linguistic diversity and vibrant cultural heritage. Language mirrors the intricate social structures and value systems in this region. For example, Hindi words \textit{Tu}, \textit{Tum}, and \textit{Aap} all mean `you', but refer to distinct levels of formality or relationship. The cultural emphasis on family structures is also clearly visible in the language usage. For example, a majority of the Indic languages have distinguishing names for paternal uncle and maternal uncle. A few of the concepts precisely elaborated in Indic languages, such as \textit{Dharma} and \textit{Karma} carry inherent philosophical ideas and are untranslatable into non-Indic languages without significant loss of meaning. The powerful connection between culture and language has also influenced political boundaries (Karnataka state for Kannada language speakers, Tamil Nadu for Tamil language speakers), and different linguistic groups share a unique cultural legacy and literary tradition. A majority of the subcontinent's history, in the form of folklore, oral epics, and proverbs, is preserved through intergenerational transmission. These oral traditions are prone to being erased when the corresponding dialect ceases to exist. 

In recent times, English serves as a lingua franca and has been intertwined such that `Indian English' incorporates the local syntax and loanwords. The common practice of code-switching (Hinglish, Kanglish) represents a modern hybrid identity that combines traditional values with global influences. It can be said that a language expresses the collective beliefs, perceptions, and value systems held by the members of the same social group, which constitute the culture of that society. With the rapid adaptation of LLMs to different language processing tasks, it is essential to inquire into the underlying cultural representation.

Foundation Models have demonstrated exceptional performance in language generation and reasoning, open-ended question answering, and multimodal understanding. While LLMs are applied across a broad range of verticals, it is necessary to understand the extensive societal consequences they entail. The paradigm shift in AI brought about by LLMs also comes with intrinsic biases such as misrepresentation, underrepresentation, and overrepresentation, and extrinsic harms that can be representational harms, abuse, and performance disparities~\cite{bommasani2022opportunitiesrisksfoundationmodels}. The downstream applications created by fine-tuning the foundation models have often displayed performance disparity in the case of minority social groups~\cite{blodgett2017racialdisparitynaturallanguage, Koenecke2020RacialASR}. 

Linguistic homogenization can be clearly identified in LLM-assisted writing. AI-enabled writing assistants have been linked to reduced linguistic diversity and alteration of individual stylistic elements while amplifying dominant characteristics or biases~\cite{sourati2025shrinking}. Repeated usage of such assistants can result in normalized grammar, simplification of dialects, and erosion of sociolinguistic markers in writing. LLMs have also been shown to erase culturally specific linguistic markers while retaining the semantic meaning, especially in professional writing~\cite{Kumar_Navneet_2026}.

LLMs also often demonstrate cultural homogenization, where they internalize dominant cultural assumptions such as the prioritization of Western values, usage of internet slang, and globally dominant political views. A cross-cultural study~\cite{10.1145/3706598.3713564} with participants from India and the United States showed that AI models such as writing assistants homogenize writing towards Western norms, decreasing subtle cultural differences. This study showed that among the two groups, Indians were more likely to write like Americans while using AI assistants, but they needed to put more effort than Americans to gain a similar productivity boost. Here, AI doesn't just change the written content but also the more internalized elements of cultural expression. Language models are also prone to follow similar writing styles and similar `neutral' tones, and can underperform~\cite {marco-etal-2024-pron} when used for creative writing tasks. Contrasted with the rich oral storytelling traditions with non-linear, layered components, LLMs follow a similar structure almost all the time. It can be said that homogenization is circular; lack of diversity in model training leads to lack of diverse outputs, which in turn guides the user expectations and standards.

The lack of cultural, epistemic, and linguistic inclusivity in LLMs is alarming. While these models are being used in different domains, including education, technology, healthcare, etc., the inherent narrative or worldview in these LMs should be understood in-depth. Especially for the Indian subcontinent, with thousands of language dialects and grouped multilingual identities, it is essential to represent the diversity in the AI ecosystem. In this section, we discuss how the language used to train the LLMs can be used to understand the embedded cultural cues. 

% We also look into future directions for working on this idea in the field of IndicNLP to promote plurality and inclusivity in foundation models.

\subsection{Worldview of LLMs}

A majority of the LLMs are trained using data extracted from conventional sources, including Wikipedia, news websites, and other publicly available datasets. For example, mBERT is trained on Wikipedia articles of the top 100 languages with the largest Wikipedia.\footnote{\hyperlink{https://github.com/google-research/bert/blob/master/multilingual.md\#data-source-and-sampling}{https://github.com/google-research/bert/blob/master/multilingual.md\#data-source-and-sampling}} These datasets sourced from such mainstream platforms are widely available as labeled gold standard corpora and are frequently used as benchmark datasets. Otherwise, extracting good-quality training corpora followed by relevant processing and labeling for the corresponding downstream task or domain is an expensive task. It requires sufficient time, computational resources, and annotators with linguistic knowledge for hand-labeling the dataset. As a result of this, a majority of the training and fine-tuning experiments are dependent on a very few public datasets that were built either by crowdsourcing or by specific vendors. This is likely to lead to standardization of algorithms due to more and more language models being trained on data from similar sources, finally leading to homogenization of the model outputs~\cite{Creel_Hellman_2022}. Language models that are trained on snapshots of the internet are more likely to mimic the views of crowd workers who were involved in the creation of data. These LMs have the least representation for a few demographic groups (for example, senior citizens, people without a college degree) that were excluded from the crowd sourcing process~\cite{santurkar2023opinionslanguagemodelsreflect}. A likely outcome of this can be that a social group with minimal representation in the dataset might be isolated in the language model system. 

The LLMs fine-tuned with human feedback are often seen demonstrating a worldview aligned to a particular group of people in response to subjective queries. A response from these models contains an embedded narrative, and it can be of great influence when used in information retrieval and decision-making tasks. For example, when it comes to political ideology, Chat GPT has expressed left-libertarian aligned views~\cite{hartmann2023politicalideologyconversationalai, HumanMotoki2023,  ceron2024beyond}. With a sufficiently large corpus of social media interactions of a political group, GPT-style models can even be fine-tuned to inquire worldviews of the particular community~\cite{jiang2022communitylmprobingpartisanworldviews}. In addition to this, the ideological stance of LLMs reflects the deep-rooted worldview of their creators. They are also influenced by the geopolitical region of their creation and the language they were created in~\cite{buyl2025largelanguagemodelsreflect}. With their increasing use in Information Retrieval (IR), it is essential to consider the origin of these models along with the worldview they reflect. 

A widely used approach for building language models for low-resource languages involves fine-tuning of multilingual foundation models. Even though this approach seems to help in overcoming the resource scarcity issue, these language models internally `think' in English while processing other languages. A research work~\cite{wendler-etal-2024-llamas} showed that English is used as the internal pivot language in multilingual models trained on unbalanced, English-dominated corpora. This work demonstrated an abstract `concept space' in the intermediate layers of the transformer model architecture, where a proximity of the embeddings to English tokens rather than the input language was observed. This finding instills English as a pivot language in a semantic sense rather than a lexical one. This kind of internal representation bias inside transformer models poses difficulty for Indic NLP by revealing the hidden English dominance. The Anglocentric bias could incline the model to certain linguistic elements, such as lexicon and grammar. Efforts such as Indic-TunedLens~\cite{indictunedlens} have proposed a framework for interpretability for Indic languages by learning shared affine transformations.

LLMs play a key role in documenting and disseminating diverse knowledge systems. They can reflect the embedded value systems and worldviews of different cultures. Due to this capability, they can be used to build local dialect datasets and community-focused language models, to preserve diverse scripts, and also for educational purposes. The current necessity is `culture sensing' where LLMs not only process or generate data, but also retain the writing style, storytelling pattern, and the innate worldview.

\subsection{Culture Sensing}

AI foundation models can operate using multiple languages. But they do not contain the lived experience of the language speakers that reflects the correlated cultural meaning. The interplay between language and culture has long been studied to understand how the perceptions, beliefs, and values of people are encoded linguistically~\cite{kramsch2014language}. Integration of AI into indigenous knowledge frameworks coming from communities needs the negotiation of needs and motivations, and establishment of relevant meanings~\cite{abundant-intelligence}. It is the need of the hour to utilize the enormous capacity of AI towards understanding and preserving the cultural and linguistic diversity. Underrepresentation of the Indic languages and the diverse cultural traditions is an understudied topic. 

Culture Sensing aims to amend the current-day foundation models based on hermeneutic reasoning. Culture Sensing approaches narrative diversity in AI using the abundant knowledge in indigenous communities that can be available mostly in the form of speech. In this section, we briefly discuss our proposed approach for culture sensing along with a few future directions.

\subsubsection{Culture Sensing for Oral Community Knowledge}\label{sec:indicasr}

% Figure is available at: https://iiitbac-my.sharepoint.com/:p:/g/personal/aparna_m_iiitb_ac_in/IQDaSoO2gDQoSLMFtOKdBvtJAcMj09DbWu1wZUvlGqa8050?e=vGhvCU

\begin{figure}
    \centering
    \includegraphics[width=\linewidth]{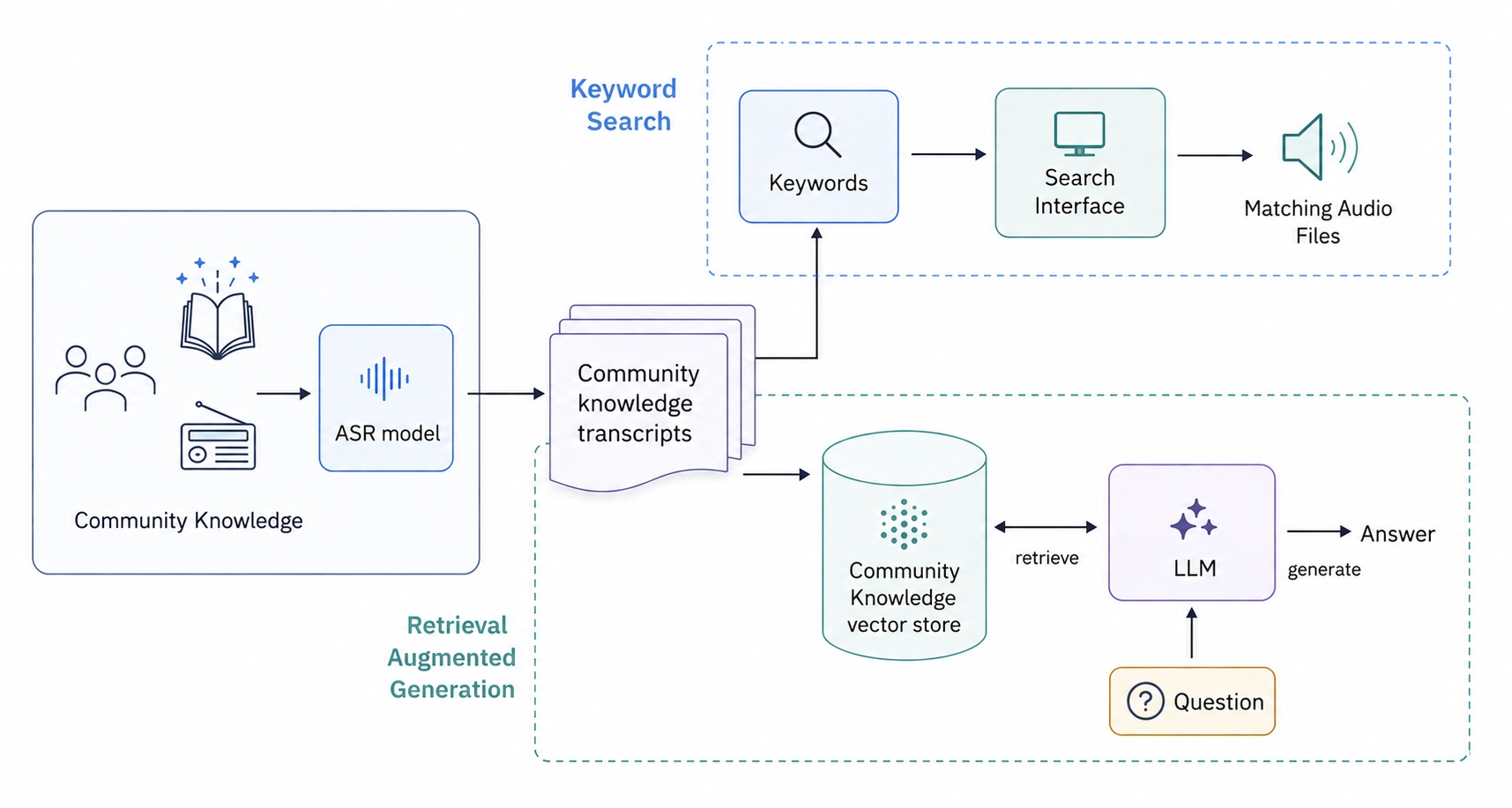}
    \caption{Reference Architecture for Culture Sensing}
    \label{fig:culturesensing}
\end{figure}

Oral data and storytelling are characteristic knowledge-sharing practices in indigenous, close-knit communities. A thorough analysis of such oral knowledge can uncover the fundamental worldview and the value system of the community in focus. Indigenous worldviews provide a holistic outlook and emphasize the relationship between humankind and nature as symbiotic. On the contrary, the mainstream or conventional worldview stems from the modern industrial advancements and is reductionist, individualistic, and places humankind in a predominant position compared to nature. A majority of the formal literature encapsulates the conventional worldview, while the native worldview is mostly present in the lived experience of the community and is context-dependent. The stark contrast between these two worldviews has resulted in the isolation of the indigenous knowledge in the mainstream systems, including AI applications. The native communities are at a disadvantage due to their limited digital presence, ultimately leading to the marginalization of their pluralistic discourse.

Bringing AI into the picture can lead to beneficial outcomes due to the trainability of the AI models. \cite{10.1007/978-3-031-58502-9_1} proposed an architecture using NLP, ASR, and Retrieval Augmented Generation (RAG) to address the problem of preserving and disseminating oral community knowledge. This architecture also aims to reveal the underlying native worldviews and also understand how they deviate from the mainstream worldview~\cite{10.1007/978-3-032-23241-0_16}. Here, an ASR pipeline is designed to convert a speech corpus to a searchable text corpus. The speech corpus is made of audio files collected from rural communities in the state of Karnataka, India. The audio files contain natural speech from the rural community members. Following this, the text corpus is utilized in multiple NLP applications such as keyword search and NER. This work also compares the community worldview with the mainstream worldview using corpora extracted from the corresponding groups. This is done using semantic analysis of embedding neighborhoods and RAG. 

A general approach for Culture Sensing is shown in Figure~\ref{fig:culturesensing}. This pipeline can be adapted for different use cases. As part of this work, we have developed two applications: \textit{Graama Kannada}~\cite{10.1145/3632410.3632483}, and \textit{Parichaya}~\cite{10.1145/3703323.3704271}. The \textit{Graama Kannada} application supports audio search and uses an audio corpus collected from the \textit{Namma Halli Radio}\footnote{\href{https://blog.janastu.org/covid-19-campaign-namma-halli-radio/}{https://blog.janastu.org/covid-19-campaign-namma-halli-radio/}} organization. This application aims to tackle two challenges: handling noisy, dialectal, and small speech corpora and maintaining low application development cost. Using a fuzzy search technique of n-gram texts, this application performs keyword search efficiently in the case of the low-resource language Kannada. The ASR model used in this work demonstrates a comparable WER to output the audio transcripts that support the keyword search functionality.

\textit{Parichaya} is an application for the management of a rural colloquial audio corpus on the topic of sandalwood cultivation. It contains two interfaces to enable users to interact with the audio content. The first interface enables browsing of the content using frequently occurring keywords and the corresponding context words, supporting a Keywords in Context (KWIC) analysis. This helps to gather an understanding of the unique aspects of information in the speech corpus. It also provides a surface-level interpretation of the corpus content. The second is a question-answering interface that provides an answer relevant to the input question along with a summary. The users can also listen to the audio fragments that contain the answers to the question. Both of these applications provides access to lesser-known corpora and enable discourse analysis to understand the crucial insights. Unlike the mainstream knowledge sources that exhibit the conventional worldview, these applications disclose some alternate narratives and diverging worldviews.

\begin{table}[h]
    \caption{Future Directions for Culture Sensing}
    \label{tab:pointers}
    \centering
    \begin{tabular}{|m{1cm}|m{12cm}|}
    \toprule
    Data  &  
            \begin{itemize}[leftmargin=*]
            \item Collect data from unconventional sources such as community radios, local community media centers, and public broadcasting 
            \item Ensure representation for multiple dialects rather than a `uniform' language style
            \item Prioritize unscripted, spontaneous, ethical, and natural data collection techniques
            \end{itemize} \\
    \midrule
    Model & 
            \begin{itemize}[leftmargin=*]
                \item Adapt existing foundation models to native language datasets by fine-tuning
                \item To counteract diglossia, extend language modeling to dialects of a language using corresponding corpora
                \item Utilize post-training techniques such as Reinforcement Learning with Human Feedback (RLHF), and inference techniques such as RAG to understand the embedded worldview
                \item Employ human-in-the-loop style verification for downstream applications
            \end{itemize}\\
    \midrule
    Users & 
            \begin{itemize}[leftmargin=*]
                \item Incentivize knowledge digitization by rural community members using easily usable platforms such as community radios
                \item More community-driven resource creation efforts
                \item Ethical data collection efforts to promote respect the cultural distinction of close-knit, sensitive communities
            \end{itemize}\\    
    \bottomrule
    \end{tabular}
\end{table}

These studies clearly showcase that the well-established, homogeneous worldview embedded in most of the datasets that are used for training the LLMs differs substantially from the numerous unique, underrepresented native knowledge systems. AI can be used to preserve such indigenous worldviews by representing the plural, heterogeneous discourses. This early study demonstrates the lack of hermeneutic diversity in foundation models. Further work in this direction can help strengthen both the cultural and linguistic representation in LLMs and the underlying narrative diversity. A few suggestions in this regard, positioned across three verticals: data, model, and users, are listed in Table~\ref{tab:pointers}.

\section{Conclusions}

The rapid growth of the Web and AI has brought out numerous applications. Especially in areas such as NLP, different paradigms have been proposed that have demonstrated exemplary performance. Deep learning has further incentivized the language modeling task by introducing foundation models with state-of-the-art performance. These developments have become both boon and bane in the case of Indic languages. On one hand, Indic NLP is becoming more efficient with time, and more LLMs for Indic languages have been coming up. However, this rapid growth has also resulted in the homogenization of hermeneutics, linguistic biases, and the risk of loss of representation of multiple worldviews.

In this paper, we present an overview of Natural Language Processing (NLP) in the context of Indic languages. We discuss the unique characteristics of these languages and how they differ from English. Additionally, we address the challenges that these distinctive features present when applying different NLP techniques to model Indic languages. Finally, we provide a comprehensive overview of the historical development of Indic NLP, highlighting the evolution of NLP techniques over time.

Ultimately, we propose Culture Sensing to enable foundation models to represent diverse and heterogeneous worldviews. Culture Sensing aims to integrate the vast amount of native knowledge available in various forms, including speech and text, into the framework of AI. This initiative is focused on ensuring that the pluralistic and vibrant worldviews are included. It addresses the urgent need for inclusivity in today's AI models. By providing a generic research direction in this area, we hope to inspire future projects that include diverse narratives from different underrepresented regions.

\section*{Ethics and Privacy Statement}
In developing the \textit{Culture Sensing} framework, we have prioritized the preservation of linguistic and hermeneutic diversity and the mitigation of algorithmic biases that would otherwise homogenize worldviews or misrepresent the lesser-known Indic languages. This work recognizes that linguistic and cultural data are important components and require high levels of stewardship. We emphasize that all data mentioned in this work are derived from publicly available, ethically sourced corpora, and this work respects the cultural nuances and intellectual authority of the represented communities. We acknowledge the potential misrepresentation of cultural narratives associated with automatic analysis and recommend human-in-the-loop validation for all downstream use cases.

\paragraph{Generative AI Usage Acknowledgment} During the preparation of this work, the authors used Generative AI tools to assist with language refinement, clear phrasing, and image editing. All content generated by these tools was thoroughly reviewed, edited, and validated by the authors.

\bibliographystyle{ACM-Reference-Format}
\bibliography{sample-base}

\appendix

\end{document}